\documentclass[journal]{vgtc}                     
\title{\name: A Visual Analytics Approach for \texorpdfstring{\\Interactive Video Programming}{}}

\author{%
  Jianben He, Xingbo Wang, Kam Kwai Wong, Xijie Huang, Changjian Chen, \\ Zixin Chen, Fengjie Wang, 
  Min Zhu, and Huamin Qu
}

\authorfooter{
  \item
  	J. He, X. Wang, KK Wong, X. Huang, Z. Chen, H. Qu are with the Hong Kong University of Science and Technology, Hong Kong, China. X. Wang is the corresponding author. E-mail: \{jhebt, xwangeg, kkwongar, xhuangbs, zchendf, huamin\}@ust.hk.
  \item
  	C. Chen is with the Tsinghua University, Beijing, China. Email: changjianchen.me@gmail.com
  \item 
        F. Wang, M. Zhu are with the Sichuang University, Chengdu, China. Email: wangfengjie@stu.scu.edu.cn, and zhumin@scu.edu.cn
}

\abstract{
Constructing supervised machine learning models for real-world video analysis require substantial labeled data, which is costly to acquire due to scarce domain expertise and laborious manual inspection.
\rw{While data programming shows promise in generating labeled data at scale with user-defined labeling functions, the high dimensional and complex temporal information in videos poses additional challenges for effectively composing and evaluating labeling functions.}
\rw{In this paper, we propose \name, a visual analytics approach to support flexible and scalable video data programming for model steering with reduced human effort.}
\rw{We first extract human-understandable events from videos using computer vision techniques and treat them as atomic components of labeling functions.}
\rw{We further propose a two-stage template mining algorithm that characterizes the sequential patterns of these events to serve as labeling function templates for efficient data labeling.}
\rw{The visual interface of \name facilitates multifaceted exploration, examination, and application of the labeling templates, allowing for effective programming of video data at scale.}
Moreover, users can monitor the impact of programming on model performance and make informed adjustments during the iterative programming process. 
We demonstrate the efficiency and effectiveness of our approach with \rc{two case studies} and expert interviews.
}


\teaser{
  \centering
  \includegraphics[width=\linewidth]{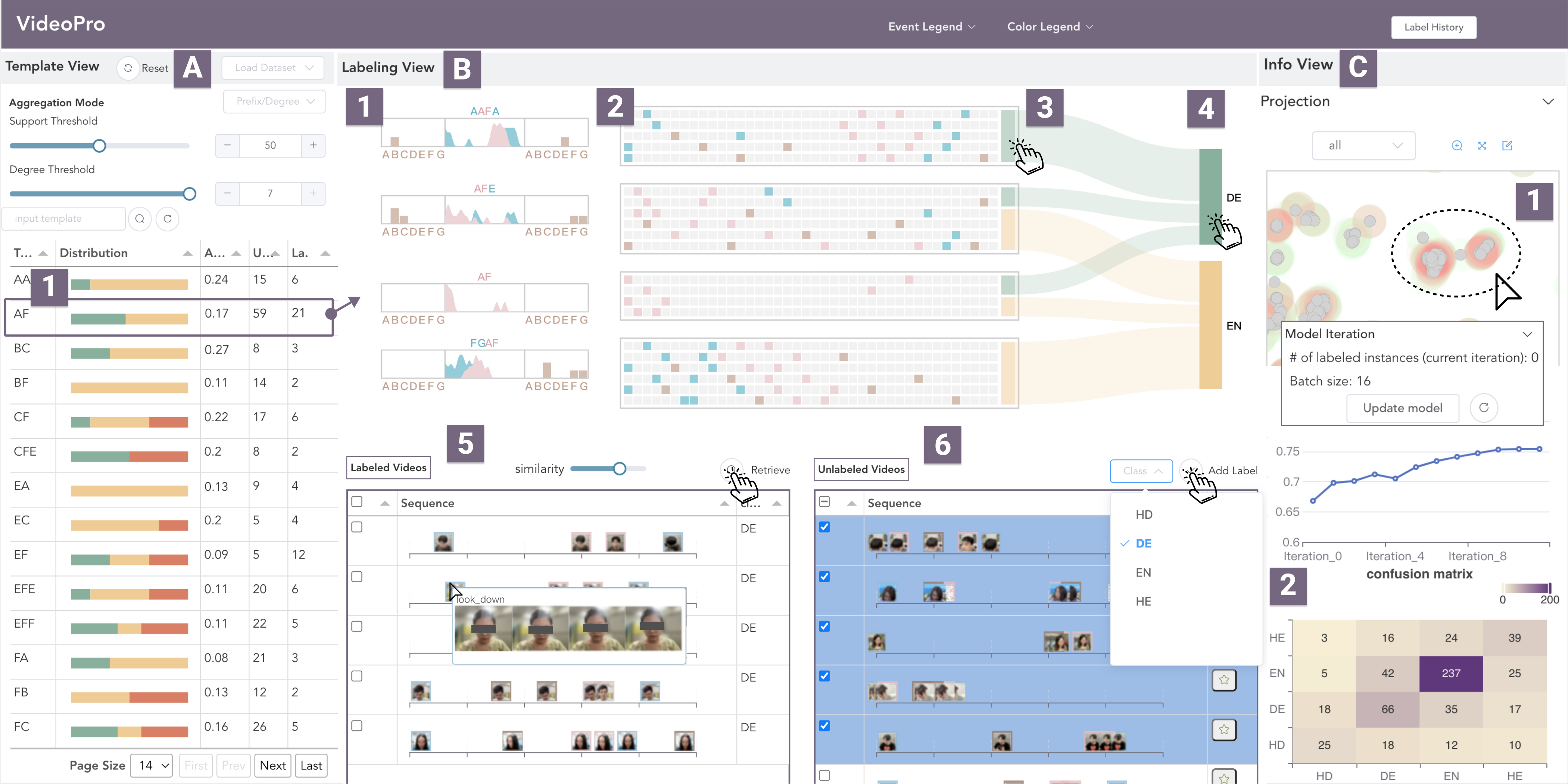}
  \caption{
The \name interface consists of three major views. The \textit{Template View} (A) offers descriptive statistics and rich interactions to facilitate multi-faceted exploration and comprehension of labeling templates. 
The \textit{Labeling View} (B) provides a summary of the nuanced event compositions within the selected template to allow effective template validation and refinement. 
\rwf{It also displays retrieved matching videos for efficient examination and at-scale programming.}
The \textit{Info View} (C) presents comprehensive information regarding data embedding distribution in latent space and the model iteration process. 
}
  \label{fig:teaser}
}

\graphicspath{{figs/}{figures/}{pictures/}{images/}{./}} 

\usepackage{booktabs}
\usepackage[svgnames,dvipsnames]{xcolor}
\usepackage{lipsum}                    
\usepackage{mwe}         
\usepackage{graphicx}

\usepackage{mathptmx}                  
\usepackage{amsmath}
\usepackage{eucal}

\usepackage{xspace,xpunctuate}
\usepackage{comment}
\usepackage{tabularx,diagbox}
\usepackage{multirow}
\usepackage[noend]{algorithmic}

\newcommand{\rc}[1]{{\color{black} #1}}
\newcommand{\rw}[1]{{\color{black} #1}}
\newcommand{\rwf}[1]{{\color{black} #1}}
\newcommand{\rcf}[1]{{\color{black} #1}}

\newcommand{\xingbo}[1]{{\color{black} #1}}

\newcommand{\ie}{\textit{i.e.},\xspace}
\newcommand{\etal}{\xspace\textit{et al.}\xspace}
\newcommand{\eg}{\textit{e.g.},\xspace}

\newcommand{\name}{\textit{VideoPro}\xspace}

\begin{document}



\firstsection{Introduction}
\maketitle
The growing prevalence of video recordings has \rw{opened up opportunities} for video analysis in numerous applications.
For instance, sports analysts analyze athletic maneuvers from recorded competitions to enhance strategic decision-making~\cite{intro_wang_2021, intro_rasipam_2023}, while scientists study videotaped experiments to identify behavioral patterns \rwf{and gather evidence to support their hypotheses~\cite{lasecki2014glance, related_video_multimodal_soure_2022}.}
Recently, deep learning models have shown remarkable potential in automatically detecting domain-specific events in videos, significantly improving analysis efficiency over manual video review~\cite{jiao2022new, vahdani2023deeplearning}.
However, building such models necessitates abundant labeled data, and the labeling process can be quite time-consuming and challenging, especially for complex video content that needs specialized domain knowledge and expertise~\cite{sun2021task}.

Data programming~\cite{li2021weakly,zhang2022survey} has emerged as a promising paradigm for reducing \rw{manual labeling efforts.}
By defining labeling functions based on their domain knowledge, users can assign weak-supervision labels to raw data for model training~\cite{ratner2016data}.
\rw{For example, for text labeling tasks, if a cluster of sentences contains similar harmful words, users can define a labeling function to assign a ``toxic'' label to the cluster~\cite{ratner2017snorkel}.}
\rw{For images, users can define a set of rules that assemble image segments (\eg head, body) to formulate new visual objects (\eg person)~\cite{hoque2022visual}.}
Nevertheless, compared with text and image data, programming video data is particularly challenging.
First, it is demanding to decompose video data into meaningful semantic units for building labeling functions. 
\rw{Videos contain segments of events that involve complex interactions among multiple objects over time.}
Particularly, the temporal context information could largely influence the semantic meaning of the video content.
For example, two cooking videos with the same food ingredients but different cooking steps can result in dishes with distinct textures and flavors.
\rw{Therefore, labeling functions need to model the variations and nuances in temporal relationships among multiple events.}
\rw{However, manually constructing such functions is challenging, given the wide range of events and their complex temporal dependencies.}

Second, evaluating, refining, and applying labeling functions for high-quality label generation and efficient model training is non-trivial.
Multiple factors, including data coverage, model performance, and semantic meanings of labeling functions, need to be considered before applying them to large unlabeled video datasets.
Furthermore, during the iterative programming process, users need to continuously monitor model performance under the impact of different labeling functions, and make corresponding refinement leveraging their domain knowledge.
Developing an effective tool to facilitate and expedite the programming process with minimal user efforts is also challenging.

To solve the above challenges, we introduce \name, a visual analytics approach that enables flexible and scalable video data programming.
\rw{Our target users are Machine Learning (ML) practitioners dealing with video datasets that have insufficient labeled examples.} 
They seek to supplement high-quality data samples for enhanced model performance of aimed tasks.
\rc{In this paper, we mainly focus on the video classification task. We also discuss how \name can be extended to support other tasks in ~\cref{sec:discussion}.}
\rwf{Drawing inspirations from the event segmentation theory~\cite{kurby2008segmentation} in cognitive science,}
we leverage Computer Vision (CV) techniques to decompose intricate video sequences into a series of human-comprehensible and semantically meaningful events.
\rw{To address the first challenge, we propose a two-stage template mining algorithm to exploit diverse event sequential patterns as templates for labeling functions.}
Regarding the second challenge, the \name interface provides carefully designed visualizations and rich interactions, allowing users to efficiently explore, validate, and refine labeling templates based on their domain knowledge.
Users can then apply the labeling functions to video data at scale and make prompt adjustments during the iterative programming process.
Our contributions are summarized as follows:
\begin{itemize}[itemsep=0pt,topsep=0pt,parsep=0pt]
\item We propose a novel approach that leverages advanced algorithms to exploit diverse event sequential patterns from videos to guide video data programming.
\item We develop a visual analytic system that provides carefully designed visualizations and rich interactions to facilitate efficient and scalable video programming.
\item We conduct \rc{two case studies} and expert interviews to validate the efficiency and effectiveness of the system.
\end{itemize}

\section{Related works}
\subsection{Interactive Data Labeling} 
\rw{A surge of research has been proposed to minimize the effort and accelerate the labeling process for supervised ML.}
These works can be categorized into model-centered and user-centered approaches~\cite{grimmeisen2022visgil}.
Model-centered approaches, exemplified by \textit{Active Learning} (AL), employ various selection strategies to prioritize the labeling of the most ``informative'' data samples, thus reducing the burden by focusing on smaller subsets of candidate instances~\cite{bernard2018towards}.
\rwf{However, AL limits users to labeling lengthy sequences of recommended instances solely determined by the selection algorithms, causing the final model performance to be heavily influenced by the selection strategies~\cite{bernard2021taxonomy}.}

Visual interactive labeling is a user-centered approach that takes advantage of users' domain expertise and visual perception to guide the selection and labeling process. 
\rw{Various visualization techniques (\eg self-organizing maps~\cite{moehrmann2011improving}, dimension reduction techniques~\cite{khayat2019vassl, liu2018crowsourcing, chen2020oodanalyzer, chen2022towards}, and thumbnail visualization~\cite{rooij2010mediatable, kurzhals2016visual}) have been employed to cluster and sort similar items for efficient labeling~\cite{yuan2021survey}.}
\rw{Recent works have incorporated more model suggestions with visualizations to further enhance labeling efficiency~\cite{chen2021interactive, yang2022diagnosing, zhang2022onelabeler}.}
For example, VIANA~\cite{sperrle2019viana} and AILA~\cite{choi2019aila} enable efficient text document labeling by visually emphasizing important text segments recommended by ML algorithms.
\rw{These mixed-initiative workflows allow users to understand and steer the models by eliciting human knowledge during the interactive labeling process~\cite{related_event_hoferlin_2012, jia2021towards}.}
Notably, PEAX~\cite{lekschas2020peax} employs the iterative labeling strategy to train classifiers for searching similar patterns in multivariate time series. 
Despite the advancements,
these approaches still face scalability challenges due to the need for manual verification of data instances one by one. We aim to address this limitation by developing a scalable solution that enables at-scale labeling and programming of video data, facilitating efficient knowledge transfer from a small set of labeled videos to a large set of unlabeled videos.

Hoque\etal~\cite{hoque2022visual} proposed the visual concept programming for image data, which is the most relevant work to ours. 
The method decomposes images into human-understandable visual concepts leveraging a pre-trained vision-language model.
Users can program these visual concepts to inject their knowledge at scale. 
However, the system primarily focuses on static spatial relationships between detected objects in images, and cannot easily generalize the resulting heuristics to temporal relationships among multiple events in videos. 
Furthermore, it relies solely on users to explore and define labeling functions and lacks prompt feedback on the impact of programming on the model performance.
To achieve a streamlined and flexible video programming workflow, we first conceptualize videos as event sequences. 
\rw{Then we propose a two-stage template mining algorithm to automatically generate labeling templates to be explored, examined, and applied, such that users can inject their knowledge via video programming in a scalable and interpretable manner.
Additionally, we offer an interim model evaluation to guide labeling focuses.}
\subsection{Visual Event Exploration in Videos}
Depending on the varying processing and target intervals, video visual analytics aims to determine the statuses in frames, detect events from scenes, and generate models for videos\cite{related_event_hoferlin_2015}.
\rw{Recent advances in CV techniques have empowered researchers to analyze videos at the frame level (\eg object detection and recognition) and study the detected objects' behaviors and interactions over extended intervals~\cite{related_event_afzal_2023}.
These behaviors and interactions are often broadly defined as ``events'' to describe the spatial and temporal dynamics within videos~\cite{related_event_schoning_2019}.}

Many Visual Analytics (VA) systems have been developed to analyze events in videos.
Li\etal~\cite{related_event_li_2021} derived anomalous events from online exam videos to support efficient proctoring.
Similarly, Tang\etal~\cite{related_event_tan_2022} detected fraudulent events in live-streaming videos with reference to streaming moderation policies.
While anomaly detection seeks to identify one anomalous event or instance as evidence, many analytical tasks require a comprehensive and multimodal context for decision-making.
As Wang\etal~\cite{related_event_mm_wang_2022} and Liang\etal~\cite{liang2023multiviz} summarized, data in different modalities can dominate, complement, or conflict with each other.
These properties have been applied to VA systems that analyze emotion~\cite{related_event_zeng_2020, related_event_maher_2022}, speech~\cite{related_event_wang_2020, related_event_wang_2022}, and body language~\cite{related_event_wu_2020, related_event_zeng_2022} in videos. 
These systems used multimodal and heterogeneous data sources to infer the actual states of events.
However, \rw{they mainly focused on one event at a time with little consideration for their temporal order, which is crucial for contextual reasoning and gaining higher-level insights}.

\rw{Parry\etal~\cite{related_event_parry_2011} identified three characteristics of events in videos, \ie \textit{hierarchy}, \textit{importance}, and \textit{state transition}.
They have inspired later research to analyze videos through the lens of event sequence understanding.}
EventAnchor~\cite{related_event_deng_2021} is developed based on the observation that badminton tactics are formulated by individual strokes, which can be detected by CV algorithms. 
From this observation, a three-level hierarchy (\ie object, event, and context) is proposed and further generalized to sports videos as the object-event-tactic framework~\cite{related_event_chen_2022} to inform the design space of augmented sports videos.
As for state transition and importance, Anchorage~\cite{related_event_wong_2023} performed event sequence analysis on customer service videos to study how different states in services affect event satisfaction ratings.
\rw{However, these works still analyze one video at a time and have low scalability.}

\rw{We aim at the data labeling scenarios, which extrapolate the event knowledge obtained from individual videos to a collection of videos.}
Over the past decade, the architecture and challenges of video labeling tools have evolved from labeling visual features~\cite{related_event_dasiopoulou_2011, related_event_hoferlin_2012} to labeling accurate event contexts~\cite{related_event_afzal_2023}.
\rw{Given the complexity of temporal information, these event contexts require additional information to assist careful human labeling for reliable knowledge injection.}
\rw{For example, users need consistency checks when coding recorded system usage videos~\cite{related_event_blascheck_2016} and temporal awareness when analyzing color usage in movies~\cite{related_event_halter_2019}.}
Similar to these video labeling tools, our approach extracts sufficient CV-based features and supports an iterative labeling process.
\rw{Furthermore, we explore the use of data programming on videos, emphasizing the events and their temporal relations to form more prominent labels.}
We propose using event sequences to distinguish and retrieve batches of videos with specific sequential patterns of interest.

\section{Requirement Analysis}
\label{sec:requirement_analysis}
\rwf{Our goal is to develop a visual analytics system that enables efficient user knowledge integration and facilitates high-quality data label generation at scale through interactive video data programming.}
The initial motivation for this research originated from our collaboration with two companies, aiming to develop high-performance models for real-world applications, including the analysis of customer and student behaviors in service and educational videos.
Considering the diverse and complex nature of events to analyze in these domain-specific videos, domain experts need to manually label the video dataset before model training. 
However, the video labeling process was time-consuming, taking several weeks even for a small-scale dataset of approximately one thousand videos, due to limited expert availability and the substantial workload involved.
Therefore, finding an efficient and scalable way to transfer domain knowledge from a small labeled video dataset to a large unlabeled one for high-quality data sample supplementation has been a persistent demand.

We worked closely with five ML experts (\textbf{E1}-\textbf{E5}, five males; three researchers, and two MLOps engineers) to understand the general needs and to derive design requirements. 
\textbf{E1}, \textbf{E2}, and \textbf{E3} are three researchers with multiple top research publications in the areas of CV and interactive ML.
\textbf{E4} and \textbf{E5} are two MLOps engineers from our collaborated company who have averaged five years of experience in developing and deploying ML models. 
Specifically, \textbf{E1} and \textbf{E3} are the co-authors of this paper.
All experts have rich experience training and utilizing ML models for video analysis. 
They highlighted that despite the availability of many public video datasets, building resilient models tailored to domain-specific tasks still necessitates significant amounts of real-world labeled data. 
Given the shortage and acquisition difficulty of such labeled data, experts expressed a desire for a tool that supports scalable knowledge transfer and efficient video programming.


The derived four design requirements are summarized as follows:
\begin{enumerate}[label=\textbf{R{\arabic*}}, nolistsep]
\item \textbf{\rc{Decompose videos with meaningful temporal event sequences}}
\rwf{All experts acknowledged the challenging and time-consuming nature of comprehending video datasets due to their large volume and rich temporal and semantic information.}
They emphasized the importance of presenting videos in a way that humans can readily understand and explore.
Particularly, experts mentioned that video contains much redundant and unimportant information. They often rely on key events to digest the entire video content, which also echoes prior research~\cite{kurby2008segmentation, related_event_parry_2011} on video understanding.
\textbf{E1} commented that ``condensing lengthy video content into a succinct event sequence enables quick grasp of the video's essence at a glance, without the need to review the entire footage.''

\item \textbf{\rc{Summarize event temporal relationships with templates from multiple facets}}
\rc{Given the large set of events in the video dataset,}
\rc{all the experts concurred that it is crucial to summarize event temporal relationships in videos with several compact templates and identify meaningful ones that can serve as labeling functions for video programming.}
\rc{Specifically, a template is a sequence of events shared by several videos, which can potentially help describe the semantics of the labels and define labeling functions for video programming.}
\rw{In addition, the experts expressed interest in exploring the templates from multiple facets, such as data coverage and model performance, to identify meaningful ones.}
For example,
\textbf{E1} prioritized templates that yield poor model performance, while \textbf{E2} focused on templates that encompass a larger number of unlabeled instances. 
\rc{\textbf{E4} showed interest in templates containing instances from a single class, indicating that ``such templates may well capture \rwf{class-specific} characteristics.''}




\item \textbf{\rc{Support efficient and scalable template-guided video data programming}}
The experts expected the system to support interactive validation and refinement of templates to achieve efficient and scalable video programming.
They pointed out that comprehending the semantic implications of templates
and verifying their correctness is crucial to ensuring \rwf{high-quality} labeling outcomes.
Moreover, the system should allow experts to refine or manually compose templates based on their domain knowledge and 
new insights that emerge during the exploration process. 
Additionally, the system should automatically retrieve the most relevant videos for programming.
This will allow users to apply selected and refined templates to program a large number of videos efficiently, 
\rc{as \textbf{E5} commented, ``it would save much effort if we could apply the knowledge to a batch of videos simultaneously.''}

\item \textbf{Reveal the effect of programming on model performance}
The experts also expressed a desire to monitor model performance changes throughout the programming process. 
They suggested that the system should provide visualizations depicting the iterative programming process to improve controllability and transparency.
They can thus gain insights into the effectiveness of selected templates and data samples as well as make corresponding adjustments in the later programming stage. 
For instance, \textbf{E3} said that when observing an unbalanced dataset distribution, he would consider adding more data samples from the minority classes to balance it.
Based on the visualized programming process, the experts can also make informed decisions about when to retrain the model and when to stop programming.
\end{enumerate}

\section{System \& Methods}

\label{sec:method}
\rwf{
In this section, we first provide an overview of the system framework and workflow. Then we illustrate the methods for video data processing, event extraction, and labeling template mining.
}

\begin{figure*}[t]
    \centering
    \includegraphics[width=\linewidth]{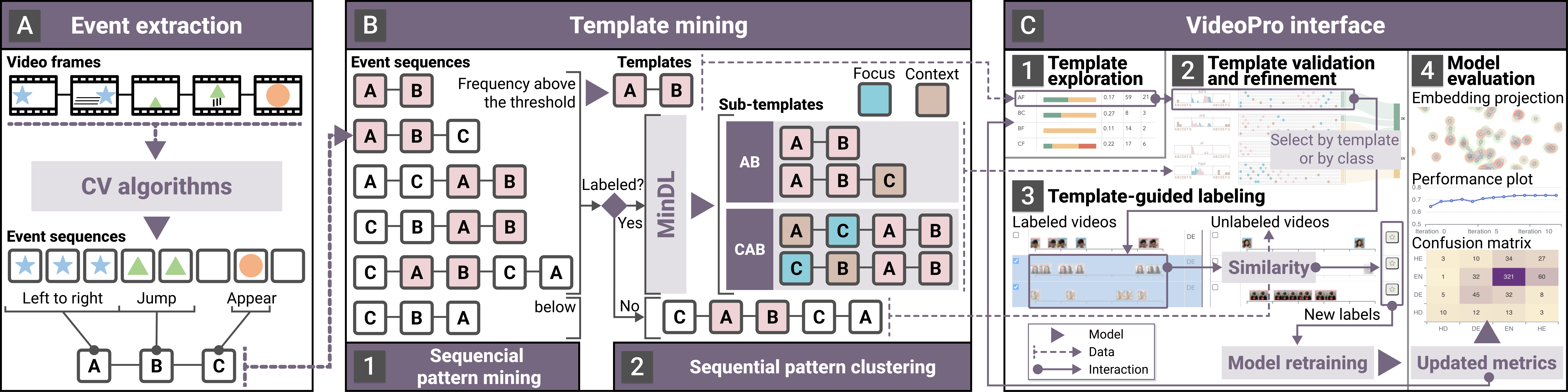}
    \caption{\rc{The system framework contains three main modules. (A) The \textit{Event Extraction} module converts input videos from the dataset into event sequences. (B) The \textit{Template Mining} module distills the event sequential patterns as templates to guide programming. (C) The \name interface supports template exploration, validation and refinement, at-scale labeling, and model evaluation for the iterative programming process.}
    }
    \label{fig:system_workflow}
\end{figure*}
\subsection{System Framework}
\Cref{fig:system_workflow} demonstrates the overarching system framework. 
The input video dataset consists of a small number of videos with ground truth labels and a substantial amount of unlabeled videos.
The \textit{Event extraction} module (\cref{fig:system_workflow}A) first abstracts the input videos as temporal sequences composed of various events (\eg wave hands) that humans can readily understand.
Subsequently, \rc{in the \textit{Template mining} module} (\cref{fig:system_workflow}B), 
\rc{a two-stage template mining algorithm is employed to extract diverse sequential patterns among events (\ie the order of event occurrence) from the collections of output video event sequences from the \textit{Event extraction} module.}
\rc{In the first stage, the sequential pattern mining algorithm (\cref{fig:system_workflow}B-1) extracts sequential patterns, which serve as potential labeling templates for programming.}
\rc{In the second stage, the MinDL algorithm (\cref{fig:system_workflow}B-2) further distinguishes and clusters the nuanced sequence variations within a template for further examination and modification.}
\rc{In the \name interface,}
Users begin by conducting a comprehensive exploration of the generated templates in the \textit{Template View} from multiple perspectives, including model accuracy and data coverage (\cref{fig:system_workflow}C-1). 
Following the selection of a template of interest, users can then efficiently validate and refine the template (\cref{fig:system_workflow}C-2), and subsequently apply the validated and refined template to label videos at scale in the \textit{Labeling View} (\cref{fig:system_workflow}C-3). The labeled instances are then forwarded to the model for retraining. 
Users can inspect and evaluate the impact of each programming iteration on the model performance in the \textit{Info View} (\cref{fig:system_workflow}C-4) and correspondingly adjust their programming strategy in the subsequent iterative programming process.

\subsection{Data and Event Extraction}
\rc{
Given input raw videos, state-of-the-art CV algorithms are leveraged to extract pre-defined events, which vary based on domain-specific requirements and expert needs. 
For instance, in application scenarios focusing on human behaviors, 
\rc{events of interest may include body movements (\eg jump and move right). These movements can be captured through analyzing position and angle changes of body parts based on heuristics and object detection models.}
} Each extracted event is represented as a tuple ($eventType$, $t\_{start}$, $t\_{end}$), where $eventType$ denotes the event type, and $t\_{start}$ and $t\_{end}$ are the timestamps of the start and end of the event.

\subsection{Template Mining}
Event sequential patterns, including the order and frequency of event occurrence, are crucial for comprehending and comparing video event sequences during programming.
Considering the diversity and complexity of event sequential patterns, we adopted a two-stage template mining algorithm~(\cref{fig:system_workflow}B)
to efficiently extract event sequential patterns and characterize the labeling templates.
\rc{
The two-stage template mining algorithm allows for scalable and generalizable analysis of large-scale datasets of varying lengths and diverse event sequential patterns.
The first frequent sequential pattern mining algorithm~\cite{wang2022seq2pat} provides a comprehensive dataset overview and avoids generating unwieldy templates that can be challenging for experts to interpret and define. It also allows users to add self-defined constraints on template compositions flexibly to accommodate their needs~\cite{wong_cohortva_2023, wong_dpviscreator_2023}. 
The MinDL algorithm~\cite{chen2018sequence, wang2022historical} in the later stage further summarizes and distinguishes nuanced sequence differences within a template to facilitate detailed validation and refinement.}

After event extraction, each video can be construed as an event sequence, denoted as an ordered event list $S = [e_{1}, e_{2}, ....e_{m}]$ where $e_{i}$ belongs to the event set $E$.
The video dataset as a whole can then be expressed as $\mathcal{S} = [S_{1}, S_{2}, ... S_{n}]$, where $n$ signifies the total number of video instances. 
A sequential pattern $P = [e_{1}, e_{2}, ... e_{|P|}]$ is a subsequence of some $S\in \mathcal{S}$ \rc{if} there exist an \rc{ordered $|P|$-tuple} $m = (m_{1}, m_{2}, ..., m_{|P|} )$ such that $S[m_{i}] = e_{i}$ \rc{for each} $e_{i} \in P$. 
\rc{For example, \rc{the sequential pattern} $P = [A, D]$ is a subsequence of $S = [A, B, D, C, D] $ with two \rc{ordered 2-tuples} (1,3) and (1,5).}
A sequential pattern is considered frequent if its occurrence exceeds a manually defined threshold.
We first employed the seq2pat algorithm~\cite{wang2022seq2pat} to extract frequent sequential patterns from the video dataset $\mathcal{S}$, which were then used as labeling templates $T = [T_1, T_2, T_3...]$.
This algorithm was chosen over other sequential pattern mining techniques due to its scalability and efficiency. 
It utilizes a multi-valued decision diagram structure~\cite{hosseininasab2019constraint} to compactly encode video sequences, enabling efficient computation for large volumes of sequences (\eg thousands) in our scenario.
Moreover, the algorithm is highly adaptable, allowing for flexible addition and revision of various constraints, such as sequential pattern length and continuity, based on user needs and task requirements.

We then implemented the MinDL algorithm~\cite{chen2018sequence, wang2022historical} to further analyze sequence nuances within a template.
This algorithm applies the minimum description length principle~\cite{grunwald2007mdl} to partition video sequence collections within the selected template into clusters and summarizes each cluster with the most ``representative'' sequential pattern, \rc{denoted as sub-template.} 
\rc{Events belonging to the selected template are denoted as core events. Events within the sub-template that are not part of the selected template are called focus events, while events outside the sub-template are referred to as context events (\cref{fig:system_workflow}B-2)}.
Every individual sequence in the cluster can be restored by editing the sub-template, including adding, deleting, or replacing events.
The total description length equals the sum of the sequential pattern length and edit length, and the optimal clustering results are obtained by minimizing the total description length $L(\mathcal{C})$:
\vspace{-0.5em}

\begin{equation}
    L(\mathcal{C})=\sum_{(P, G) \in \mathcal{C}} \operatorname{len}(P)+\alpha \sum_{(P, G) \in \mathcal{C}} \sum_{s \in G}\|edits(s, P)\|+\lambda\|\mathcal{C}\|
\end{equation}

\vspace{-0.5em}

Here, $\mathcal{C}$ denotes the collection of video sequences in a template. $s$ represents the individual video event sequence. The divided sequence clusters are denoted as $\mathcal{C}=\left\{\left(P_1, G_1\right),\left(P_2, G_2\right), \ldots,\left(P_n, G_n\right)\right\}$ where $P_i$ and $G_i$ are the representative sequential pattern and sequence collection of the $i^{th}$ cluster. 
The parameters $\alpha$ and $\lambda$ respectively control the information loss importance and the number of clusters. 
Based on our experiment results, we found that setting $\alpha$ as 0.8 and $\lambda$ as 0 can yield a satisfactory summary for our dataset. 
We adopted a similar Locality Sensitive Hashing (LSH) strategy~\cite{chen2018sequence, wang2022historical} to speed up the computation. 
We also modified the original algorithm to adapt to our problem. Specifically, the computed representative sequential patterns of all clusters must include the original template for effective understanding and comparison. 
The MinDL algorithm excels in partitioning sequences into meaningful clusters based on temporal similarity and identifying representative sequential patterns to provide an informative summary. This is particularly useful for users to compare and understand different video sequence clusters for further labeling template validation and refinement in our scenarios.

\section{User Interface}
\label{sec:system}
The \name interface consists of three coordinated views (\cref{fig:teaser}) to support flexible and smooth programming experience. In this section, we introduce the visual design of each view and the interactions connecting them in detail.
The \name adopted a unified color and event encoding scheme that is displayed at the top of the system interface. 
In consideration of scalability and generalizability, we use alphabets instead of icons or colors to encode individual events.
\subsection{Template View}
\rw{The \textit{Template View} (\cref{fig:teaser}A) summarizes the frequent and influential labeling templates in an organized table.}
It facilitates multi-faceted template exploration and comprehension (\textbf{R1, R2}).

The first column in the \textit{Template View} records the template name, which indicates the summarized event sequential patterns.
The second column uses a stacked bar chart to encode the class distribution of labeled video instances included in the corresponding template. 
The length of the bar chart encodes the video instance number, while the color encodes the class type. 
\rw{Hovering over the bars of different colors shows each class's exact number of labeled video instances, providing a clear understanding of the class distribution within the template.}
\rw{The bar charts will be updated after each labeling round.
Newly labeled instances are visually distinguished from previously labeled ones using the corresponding class color and a check texture.}
The third and fourth columns respectively display the overall prediction accuracy of labeled video instances and the number of unlabeled instances within the template, which will also be updated after each labeling round.

A control panel on the top of the template table offers multiple interaction options, where users can choose to aggregate templates in different ways, including by prefix, by degree \rc{(\ie template length)}, and by set \rc{(\ie event collections in template)}. 
By default, templates are aggregated by prefix. 
Users can expand templates for further exploration by clicking the ``+'' symbol.
Users can customize the \textit{Template View} based on their specific needs by setting frequency and degree threshold to filter templates. 
They can also sort the templates by multiple predefined metrics, including overall prediction accuracy, unlabeled video instance number, and label purity in ascending or descending order.
\rw{In addition, users can manually input and search for templates based on their domain knowledge in the search box above the table.}

\subsection{Labeling View}
Upon selecting a template in the \textit{Template View}, users can validate and refine the selected template, as well as examine the videos that match the template for scalable labeling in the \textit{Labeling View} (\textbf{R1, R3}).

\begin{figure}[ht]
    \centering
    \includegraphics[width=\linewidth]{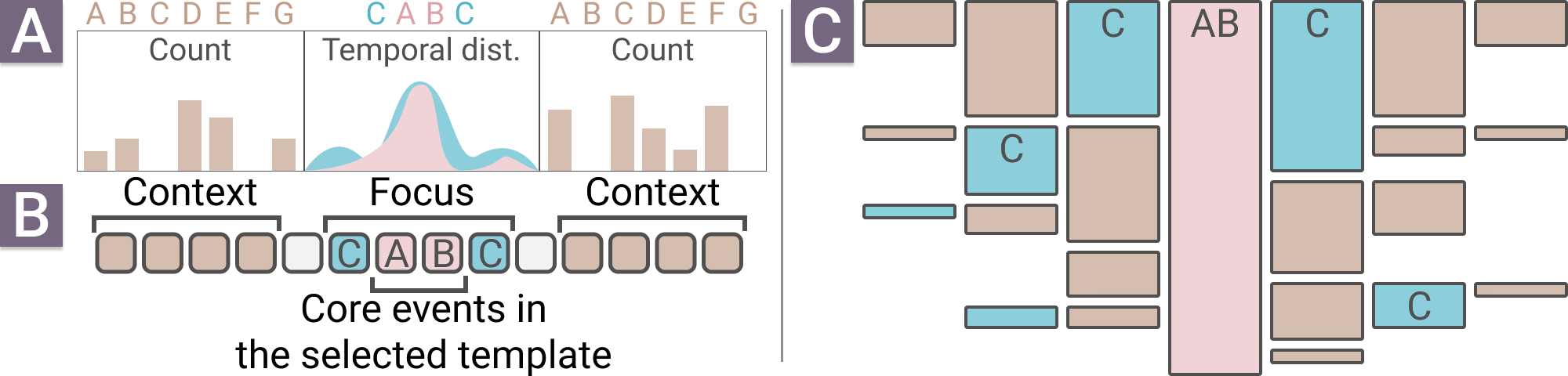}
    \caption{\rc{The design for event sequential pattern summarization. (A) our current design based on the periphery plots~\cite{periphery_plots}.
    (B) an illustration of the original event sequence. (C) an icicle plot alternative design.}}
    \label{fig:alternative}
\end{figure}

\rc{The upper part of the view (\cref{fig:teaser}B) consists of three parts from left to right: the summary figures, the cluster heatmaps, and the connected Sankey diagrams.}
The summary figures (\cref{fig:teaser}B-1 \rc{and \cref{fig:alternative}A), inspired by the periphery plots~\cite{periphery_plots},} provide an overview of the temporal event distributions within the corresponding clusters.
The middle \rc{stacked line charts} depict the aggregated temporal distribution of the sub-template events across the entire video clusters, while the histograms on either side illustrate the frequency of context events occurring before and after the sub-template events. 
This design enables users to compare the event temporal distribution of sub-templates and observe the differences in contextual events between and within clusters.

The middle cluster heatmaps show the temporal distribution of the labeled videos belonging to the clusters. 
Each row represents an individual video sequence, and each grid represents a fixed time interval (\cref{fig:teaser}B-2).
For example, if one video is 10 seconds long and there are 10 grids, then each grid represents 1 second time interval.
To facilitate cross-video temporal comparisons, the time duration of all video sequences is normalized so that they contain the same number of grids. 
Videos belonging to the same cluster are vertically stacked together, with larger clusters having larger heights. 
\rc{The color of each grid indicates the types of events occurring during the corresponding time interval, including core events from the selected template, the focus events in the sub-template, and other context events.}
Users can hover over the grid to inspect the specific event.

Furthermore, a Sankey diagram-based design (\cref{fig:teaser}B-(3-4)) is adopted to visualize the label distribution across different clusters.
The colored bar at the end of each video sequence indicates its label class.
Therefore, the height of the colored bars at the end of each cluster (\cref{fig:teaser}B-3) reflects the number of video instances belonging to the corresponding class in the cluster.
\rw{The rightmost colored rectangles (\cref{fig:teaser}B-4) represent corresponding classes and are linked with their contained video instances (\ie the colored bars) through flows of different widths.}
The width of the flows equals the bar height, thereby encoding the total number of video instances for each class.
\rw{Hovering over a rectangle will highlight all associated flows.}
\rw{Additionally, users can click on each rectangle to stack videos of the same class together for efficient comparison.}
Additionally, users can select a group of videos by clicking on the corresponding colored bar. Then the original video keyframe sequences of the selected group will be displayed below.

\textbf{Alternative design} 
\rc{
Another candidate design based on the icicle plot (\cref{fig:alternative}C) was considered to visualize the sub-templates. 
By accumulating all event sequences in the sub-template as one, the bars' height encodes the event occurrence in order.
These events are aligned by the selected template, with the before-and-after events on the sides.
However, the design does not scale when the sequences are long (\ie too many layers on both sides).
It could also mislead users that the left and right sub-sequences belong to some actual sequences if they share the same horizontal positions.
For example, the CABC in the middle of \cref{fig:alternative}C might not exist in the cluster.
To fix these critical flaws, we adopt the current design that shows the overview and details separately.
}

The lower part of the \textit{Labeling View} presents the original video content to facilitate quick examination and at-scale labeling (\cref{fig:teaser}B-(5-6)). 
The lists of labeled and unlabeled videos are displayed on the left and right sides respectively, enabling straightforward comparison. 
Unlabeled videos are ranked based on their similarities to labeled videos by default.
\xingbo{The similarity ($Sim_{total}$) between an unlabeled video and a labeled video is modeled by a linear combination of similarities of both event sequence ($Sim_E$) and video embedding ($Sim_V$): $Sim_{total} = w \cdot Sim_E + (1-w) \cdot Sim_V$. The $Sim_E$ is measured using editing distance to compare the discrete event sequences of the two videos, while $Sim_V$ is measured using cosine similarity to compare their video embeddings. The weight factor $w$ balances the assessment of patterns of interest ($Sim_E$) and overall video visual similarities ($Sim_V$).}
Users can adjust the similarity slider to control video similarity, and retrieve similar unlabeled videos by selecting corresponding labeled videos and clicking the retrieval button.
In the video list, each row represents a single video. Each event within the video sequence is succinctly summarized using the extracted keyframe. 
\rw{The position of the keyframe on the horizontal timeline encodes the event's occurrence time, while the border color indicates the event type (core, focus, or context event).}
Users can hover over the keyframe to browse the complete frame sequences of the event.
They can also click on the row to play the original videos for detailed inspection and bookmarking.
This design allows efficient video content digestion and intuitive comparison of the event temporal distribution.
Users can apply a label to multiple selected videos at once by checking corresponding selection boxes in an efficient and user-friendly manner.
Users can also check the labeling history and resolve labeling conflicts in the upper-left labeling history panel.

\subsection{Info View}
The \textit{Info View} (\cref{fig:teaser}C) provides comprehensive information about data embedding distribution and model iterations (\textbf{R4}).

The \textit{Projection} design (\cref{fig:teaser}C-1) provides an overview of data instances by displaying their label status and latent space similarity. High-dimensional latent embeddings are projected onto a 2D plane using the UMAP algorithm~\cite{mcinnes2020umap}, resulting in data instances with similar embeddings positioned close to each other. 
Labeled and unlabeled data instances are differentiated using two distinct colors.
Users can select to view all data instances or focus on partial(\ie labeled or unlabeled) data instances from the top menu.
A heatmap is added in the background to encode the prediction error of data instances, with the redder shades indicating higher prediction errors. 

The \textit{Model Iteration} part (\cref{fig:teaser}C-2) serves to update users about the impact of each iteration of programming on the model training progress.
It includes an overall model accuracy line chart and a confusion matrix for model performance evaluation.
The line chart shows how overall model accuracy changes with the number of labeled instances.
The x-axis indicates the number of labeled instances while the y-axis indicates model accuracy.
To ensure computation efficiency, retraining occurs when the number of newly labeled instances reaches the batch threshold, and the line chart will be updated accordingly.
The confusion matrix, color-coded with a sequential colormap, shows the proportion of correctly classified video instances per class. 
The rows and columns represent ground truth classes and predicted classes respectively. 
Users can analyze classifier performance across classes, guiding template selection and data supplementation in subsequent programming iterations.

\subsection{Cross-view Interactions}
The \name system offers diverse interactions for seamless coordination of different views with on-demand access to details.

\textbf{Clicking}
Users can double-click on a specific template to inspect labeled and unlabeled video instances belonging to the template in the \textit{Labeling View} and highlight them in the \textit{Info View} projection plane. The reset buttons can be used to undo any operations. 


\textbf{Lasso and zooming}
Users can leverage the lasso and zoom interactions in the \textit{Info View} projection to inspect and select instance groups of interest. The corresponding templates will be computed and updated in the \textit{Template View}.

\section{Evaluation}
\label{sec:evaluation}
In this section, we demonstrate the efficiency and effectiveness of our system through two case studies and domain expert feedback.
The first case study is conducted on a real-world online education video dataset provided by our collaborated speaking training company. 
This dataset was used to build a robust classification model for assessing students' engagement levels in online classes, as no related public datasets or models were available. 
The second case study is performed on the UCF101 dataset~\cite{soomro2012ucf101}, a representative public action recognition dataset, for the action classification task.
The primary goal of these two case studies is to facilitate experts in efficiently supplementing high-quality data samples using \name, achieving satisfactory model performance with minimal effort.

\subsection{Case One: Engagement Classification}
\label{sec:case_one}
We invited expert \textbf{E1} to conduct the case study.
As a member of the collaborated project, \textbf{E1} has been responsible for developing a classification model on this dataset and involved in the prototype design of our system. 
He thus has a good understanding of the task, dataset, workflow, and system design.

\textbf{Dataset} 
The whole video dataset contains 5,788 videos in total, including 1,774 videos with four-class ground-truth labels and 4,014 videos without labels.
\rc{For the labeled videos, 
the label falls into four classes: \textit{Highly Disengaged (HD)}, \textit{Disengaged (DE)}, \textit{Engaged (EN)}, and \textit{Highly Engaged (HE)}.}
\rc{This classification scheme is established according to the experts' requirements and previous work practices~\cite{bagher-2018-multimodal, tsai-2019-multimodal}.}
\rcf{The class distribution of the labeled videos is as follows: \textit{HD} (8.68\%), \textit{DE} (23.96\%), \textit{EN} (52.03\%), and \textit{HE} (15.33\%).}
Following MS COCO~\cite{lin2014microsoft}, we further split the videos with ground-truth labels at the proportion around 2:1 into the training and test sets. In the splitting process, we maintain the label distribution of four classes to be the same in both the training and testing sets. The training set contains 1,182 videos, and the test set contains 592 videos.

\textbf{Initial Setting}
To understand typical events for assessing student engagement levels, we interviewed three experienced teachers from our collaborating company. These teachers, with rich domain knowledge, are also responsible for labeling a small subset of the dataset.
Ultimately, the consolidated event set $E$ consisted of seven types of events: active hand movement, look away, look center, smile, look down, move away from the screen, and move close to the screen.
We leveraged several state-of-the-art CV techniques~\cite{baltruvsaitis2016openface, hand_detection, smile_detection} to extract these representative events from videos.
\rc{Initially, we trained a baseline classifier that integrated spatiotemporal features extracted by I3D~\cite{carreira2017quo}, a state-of-the-art pre-trained model, and event features represented by one-hot encoding.}
We use the accuracy for each class and the overall F1 score to evaluate the model performance.
It achieved an overall F1 score of \rc{66.78\%} on the test set, where its performance is recorded in the first row of \cref{tab: performance improvement}.

\textbf{Iteration One: Distinguish between \textit{DE} class and \textit{EN} class}
After the initial round of training, \textbf{E1} observed that the model performance was unsatisfactory in distinguishing between the \textit{DE} class and \textit{EN} class.
He suspected that the model struggled to effectively differentiate some videos within these two classes that share similar \rw{event sequential patterns}. 
\rw{To address this issue, \textbf{E1} aimed to identify common templates with low accuracy that were shared between the \textit{DE} class and \textit{EN} class.}

While examining the projection in the \textit{Info View} (\cref{fig:teaser}C-1), \textbf{E1} identified a group of video embeddings highlighted with a red-colored background, indicating high errors.
To further investigate these videos, \textbf{E1} utilized the lasso tool to select them, and the corresponding templates that characterized these videos were shown in the \textit{Template View} (\textbf{R2}).
\textbf{E1} observed that the template ``AF'' exclusively contained videos from the \textit{DE} and \textit{EN} classes, as indicated by the two-color distribution bar chart (\cref{fig:teaser}A-1). This template also contained a relatively large number of labeled and unlabeled videos. Therefore, he decided to further investigate the ``AF'' template by double-clicking on it to examine its contained videos in the \textit{Labeling View}.

\begin{figure}[ht]
    \centering
    \includegraphics[width=0.9\linewidth]{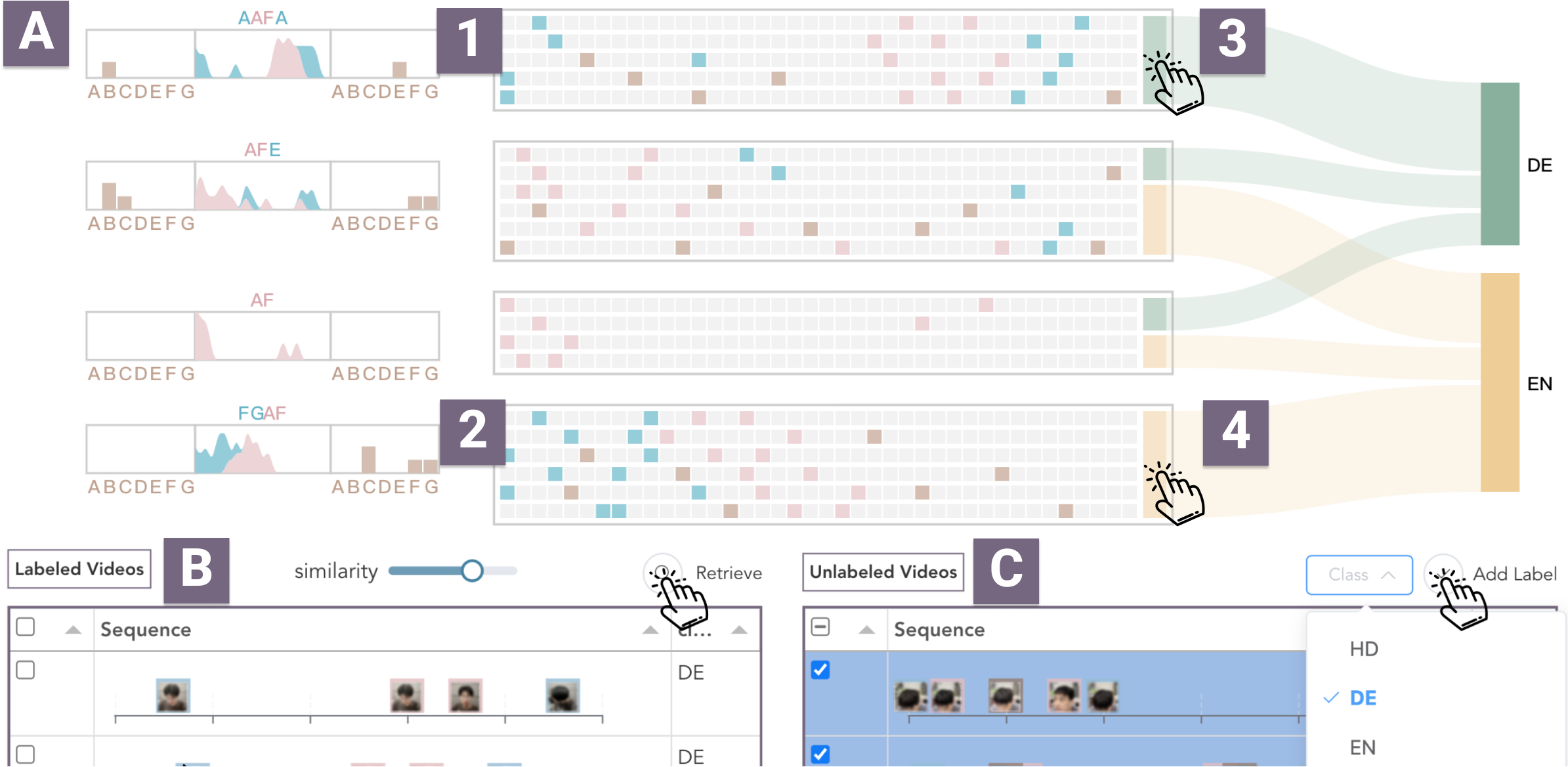}
    \caption{\rcf{The \textit{Label View} in the case one.(A) the sub-templates within the selected ``AF'' template. (B) the corresponding labeled videos and retrieved unlabeled videos when clicking on the green colored bars.}}
    \label{fig:caseone}
\end{figure}

Upon observing the flows between these clusters and their corresponding classes in the \textit{Labeling View}, \textbf{E1} found that two clusters exclusively contained videos from the \textit{DE} class (\cref{fig:caseone}A-1) and \textit{EN} class (\cref{fig:caseone}A-2) respectively. 
\textbf{E1} also inspected the representative sequential patterns and event distributions in the left summary figure to better understand the relationship between the sequence orders within the same template and class results (\textbf{R3}).
\rw{The \textit{DE} cluster was characterized by the sequence ``AAFA'', while the \textit{EN} cluster exhibited the sequential pattern ``FGAF''.}
\rw{Through analyzing the event distribution histogram,}
\textbf{E1} also noticed that the \textit{DE} cluster had a higher \rw{occurrence} of events involving moving far away and looking away, while the \textit{EN} cluster had a higher occurrence of events such as looking center and moving closer to the screen.

\rw{After examining the labeled videos by clicking on the colored bars (\cref{fig:caseone}A-(3-4)),} 
\textbf{E1} observed that participants classified as \textit{DE} frequently looked down, appeared preoccupied with their own work, and only occasionally \rw{directed their attention to the center of the screen}. 
In contrast, participants classified as \textit{EN} listened attentively with their eyes focused on the center of the screen most of the time, looked down for a short time, and exhibited positive behaviors like smiling.
These observations led \textbf{E1} to conclude that these two summarized sequential patterns effectively characterized the \textit{DE} and \textit{EN} classes.
Consequently, \textbf{E1} felt confident in using these two refined templates for data supplementation to highlight the differences between the \textit{DE} and \textit{EN} classes.
\rw{By clicking on the Retrieve button (\cref{fig:caseone}B), \textbf{E1} obtained the unlabeled videos exhibiting similar patterns for efficient labeling (\textbf{R3}).}
Through browsing the keyframes and their border colors, \textbf{E1} quickly identified the videos that closely matched the two representative patterns (\textbf{R1}). 
He then selected these videos by checking the selection boxes and applying the corresponding class label to them all at once (\cref{fig:caseone}C).



\textbf{E1} then initiated model retraining in the \textit{Info View}.
The results of this iteration are shown in the second row of ~\cref{tab: performance improvement}. 
Compared with the initial baseline, the performance of the \textit{DE} and \textit{EN} classes improved \rc{+3.86\%} and \rc{+2.29\%} respectively.
\rw{This result indicated the effective utilization of the acquired knowledge about the distinction between classes in supervising model training, which was achieved by supplementing high-quality labels using refined templates.}
Meanwhile, \textbf{E1} noticed that the performance of the \textit{HD} and \textit{HE} classes significantly dropped.
Considering the absence of supervision for the other two classes in this round, he thought this outcome was reasonable. 
As a result, \textbf{E1} planned to augment the model's understanding of the other two classes in the next iteration (\textbf{R4}).

\begin{figure}[ht]
    \centering
    \includegraphics[width=.7\linewidth]{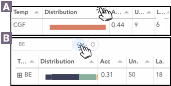}
    \caption{\rc{The representative templates for \textit{HE} and \textit{HD} class. (A) the template only contains videos belonging to \textit{HE} class. (B) the searched template based on domain knowledge.}}
    \label{fig: balance data distribution}
\end{figure}

\textbf{Iteration Two: Balance dataset distribution}
After examining the labeled data distributions, \textbf{E1} noticed an imbalance across the four classes, where the \textit{HD} and \textit{HE} classes had only a few video samples. 
To improve the model's robustness and stability, \textbf{E1} decided to use \name to supplement more samples from the two minority classes.

\rw{To identify representative templates for quick labeling of the \textit{HE} class (\textbf{R2}), \textbf{E1} sorted the templates in the \textit{Template View} based on the descending purity value by clicking the distribution column. }
\rwf{The top-ranked template ``CGF'' displayed a red distribution bar, indicating that all labeled videos in the template belonged to the \textit{HE} class (\cref{fig: balance data distribution}A).}
After randomly selecting labeled videos and browsing the original videos in the \textit{Labeling View}, \textbf{E1} observed that this behavior sequence frequently occurred when students were deeply engaged in the teaching content. They tended to approach the screen, respond with smiles, and maintain focused attention on the screen (\textbf{R1}).
As the unlabeled videos were ranked based on similarity, \textbf{E1} directly checked the last retrieved unlabeled video by hovering over its keyframes for quick validation. 
He found that the scenarios in this video aligned well with the labeled ones, which increased his confidence in the template. 
\rwf{Therefore, \textbf{E1} labeled these retrieved unlabeled videos as \textit{HE} class at scale (\textbf{R3}).}

In the sorted templates based on purity value, \textbf{E1} failed to find a template with a pure blue bar in the distribution column, indicating the absence of exclusively labeled videos for the \textit{HD} class.
This observation reinforced the need to supplement more data from this class to achieve dataset balance. 
Drawing on his general knowledge and previous discussions with domain experts, \textbf{E1} recalled that the pattern of moving away from the screen and then consistently looking away is often associated with high disengagement.
Thus he searched the corresponding template ``BE'' directly in the search box in the \textit{Template View} (\cref{fig: balance data distribution}B).
As a result, the template ``BE'' appeared at the top with a distribution bar that has a large portion of blue.
Most of the retrieved unlabeled videos had a good match where he applied the \textit{HD} label in a similar manner. After adding more samples to these two minority classes following a similar process, \textbf{E1} proceeded to send the newly supplemented samples for model retraining.

The outcomes of the second round of iteration are shown in the third row of ~\cref{tab: performance improvement}. It is evident that following the introduction of additional knowledge and supplementation of data for the two underrepresented classes, the model exhibited significant improvements in its performance in these classes (\rc{+7.24\%} and \rc{+5.65\%} for \textit{HD} class and \textit{HE} class respectively). Moreover, the overall model performance has also been improved. 
After \rc{10} programming iterations, \textbf{E1} noticed that the overall accuracy ceased to increase and instead stabilized at around \rc{75.4\%}. 
This result also satisfied the project objective of achieving an overall classification accuracy above 70\%.
The final overall accuracy and each class all improved compared with the initial baseline.
Consequently, \textbf{E1} is satisfied with this programming result and decided to stop programming  (\textbf{R4}).

\begin{table}[htb]
\centering
\caption{Performance improvement using \name on the engagement classification task, measured by the accuracy of each class and overall F1 score (larger is better).}
\newcolumntype{A}{>{\hsize=.375\hsize}X}
\newcolumntype{B}{>{\hsize=.1125\hsize}X}
\newcolumntype{C}{>{\hsize=.175\hsize}X}
\resizebox{0.9\linewidth}{!}{
\begin{tabularx}{\columnwidth}{A|BBBBC}
\toprule
\diagbox[innerwidth=7.25em]{\textbf{Setting}}{\textbf{Results(\%)}}  & \textbf{HD} &\textbf{DE} & \textbf{EN}& \textbf{HE}& \textbf{F1 score}\\ \midrule
Baseline   &    47.62   &   49.31   &  76.82  &   43.42   &  66.78       \\
Iteration One &    40.21  &    53.17      &     79.11       &    35.78     &    69.82    \\
Iteration Two &    47.45  &    49.04      &   78.29     &     41.43  &      70.12  \\
Iteration Ten &    55.36  &     54.07      &    80.14    &      49.64         &   75.43      \\
\bottomrule
\end{tabularx}
}
\label{tab: performance improvement}
\end{table}

\textbf{Post Analysis}
Following the case study, a quantitative experiment was conducted to compare the labeling efficiency of \name with an active learning-based labeling baseline approach.
The baseline approach utilized the uncertainty-based strategy, a widely adopted technique in active learning~\cite{settles1994active}, that selects the most uncertain videos for labeling at each time. 
The experiment results are summarized in ~\cref{tab: label_accuracy}.
It showed that the active learning-based approach required labeling \rc{2081} video samples to achieve an overall accuracy of \rc{75.38\%}. In contrast, \name enabled the expert to label \rc{10} iterations and \rc{452} samples in total, achieving an overall accuracy of \rc{75.43\%}.

Furthermore, we compared the time cost of the two approaches. 
The time cost for the baseline approach was estimated based on the average time needed for labeling a single video by domain experts (i.e., teachers) using the labeling tool provided by the collaborated company. 
The average labeling time was half a minute per video as recorded, resulting in a total time cost of \rc{17.3} hours.
In comparison, \name recorded the total operation time, where the expert took \rc{1} hour to finish all the labeling. 
The experiment results show that \name incurs lower labeling and time cost than the baseline approach to attain comparable levels of accuracy.
It demonstrates that \name significantly improves labeling efficiency.

\begin{table}[htb]

\centering
\caption{The number of labeled samples and time cost comparison between the active learning-based approach and \name.}
\newcolumntype{B}{>{\centering\arraybackslash \hsize=.3\hsize}X}
\newcolumntype{D}{>{\hsize=.2\hsize}X}
\newcolumntype{C}{>{\centering\arraybackslash \hsize=.175\hsize}X}
\newcolumntype{A}{>{\hsize=.2\hsize}X}
\renewcommand{\arraystretch}{1.5}
\resizebox{0.9\linewidth}{!}{
\begin{tabularx}{\columnwidth}{c|c|B|c|C}
\toprule
\textbf{Dataset} &\textbf{Method }  & \textbf{\# of labeled samples} &\textbf{time cost} & \textbf{F1 score(\%)}\\ \midrule
\multirow{2}{*}{\textbf{Engagement}} & Baseline &   2081 & 17.3h & 75.38 
 \\ \cline{2-5} 
& VideoPro &  452 & 1.0h & 75.43 \\ \midrule
\multirow{2}{*}{\textbf{UCF101}} & Baseline &  496 & 0.8h & 93.92 
 \\ \cline{2-5} &
 VideoPro &  304 & 0.5h & 93.98 \\
 
\bottomrule
\end{tabularx}}
\label{tab: label_accuracy}

\end{table}


\subsection{Case Two: Action Recognition on UCF101 dataset}
\rc{To further validate the effectiveness and generalizability of \name, we extended our system for a more general action recognition task.
We invited \textbf{E6}, a sports analytics researcher who has published multiple articles about sports-related labeling and analytics tools, to conduct this case study.
He has rich experience building sports analytics models and extensive knowledge in the sports and exercise domain.}

\rc{\textbf{Initial Setting}
For the system demonstration, the expert selected 10 sports-related action classes that he is familiar with from the UCF101 dataset. These action classes include \textit{Archery \rcf{(10.29\%)}, CleanAndJerk \rcf{(7.35\%)}, Basketball Shooting \rcf{(10.29\%)}, High Jump \rcf{(11.76\%)}, Javelin Throw \rcf{(14.71\%)}, Tennis Swing \rcf{(7.35\%)}, PullUps \rcf{(4.41\%)}, PushUps \rcf{(11.76\%)}, Lunges \rcf{(8.82\%)}, Body Weight Squats \rcf{(13.24\%)}}.
\rcf{The number in the bracket indicates the corresponding class distribution in the dataset.}
To identify the fine-grained semantic events associated with these activities, we conducted interviews with \textbf{E6} and his colleagues, and extensively reviewed relevant literature in the field.
Drawing from experts' insights and borrowing concepts from relevant sports biomechanics research\cite{abernethy2013biophysical}, we defined a set of fine-grained semantic events that include arm flexion (A), arm extension (B), arm abduction (C), arm adduction (D), leg flexion (E), leg extension (F), leg abduction (G), and leg adduction (H), body elevation (I), and body depression (J). 
To detect these events, we first adopted the advanced pose detection model~\cite{cao2019openpose} for body keypoint and part detection.
We then utlized rule-based heuristics\cite{hamill2006biomechanical} to detect these events. 
Specifically, we calculated the displacement of body parts along and perpendicular to the body's midline to detect abduction/adduction and elevation/depression events. 
Additionally, we measure angle changes between body parts to detect flexion/extension events.

We followed the original training-test split of the UCF101 dataset on the selected 10 classes. The constructed 10-class dataset thus contains 1,016 videos in total, with 733 videos in the training set and 283 videos in the test set.
We further split the training set into the labeled dataset with 68 videos and the unlabeled dataset with 665 videos to simulate the scenarios with very few labeled videos at the beginning. 
The label distribution in the original dataset is preserved during the splitting process.
We adopt the state-of-the-art uniFormer backbone~\cite{li2022uniformer} to train a baseline classification model on the constructed labeled dataset (with 68 videos). It achieves an overall F1 score of 82.69\% on the test set. 
}


\rc{\textbf{Programming Process}
After analyzing the performance of the baseline model on the test set, \textbf{E6} observed that the model performed poorly on the \textit{High Jump} and \textit{Javelin Throw} classes. 
The confusion matrix further indicated the model's inability to distinguish between these two classes. 
Therefore, \textbf{E6} decided to supplement more labels for these two classes. 
Looking at the \textit{Template View} sorted by prediction accuracy, \textbf{E6} discovered the template ``EFEFEF'' (\cref{fig: ucf_labelview}A), which indicates repetitive leg flexion and extension movements, contained a large portion of videos from these two classes (\textbf{R2}).
Drawing from domain knowledge, \textbf{E6} pointed out the distinct stages within the \textit{High Jump} and \textit{Javelin Throw} activities. The \textit{High Jump} can be roughly divided into approach, takeoff, and landing stages, while the \textit{Javelin Throw} activity involves stages such as approach, windup, and release.
These two actions shared common initial event sequences involving repetitive leg movements to generate momentum during the approach stage.
Recognizing the potential value of this template in representing these two classes, \textbf{E6} proceeded to explore its contained sub-templates in the \textit{Labeling View} by clicking on the template.

\begin{figure}[ht]
    \centering
    \includegraphics[width=\linewidth]{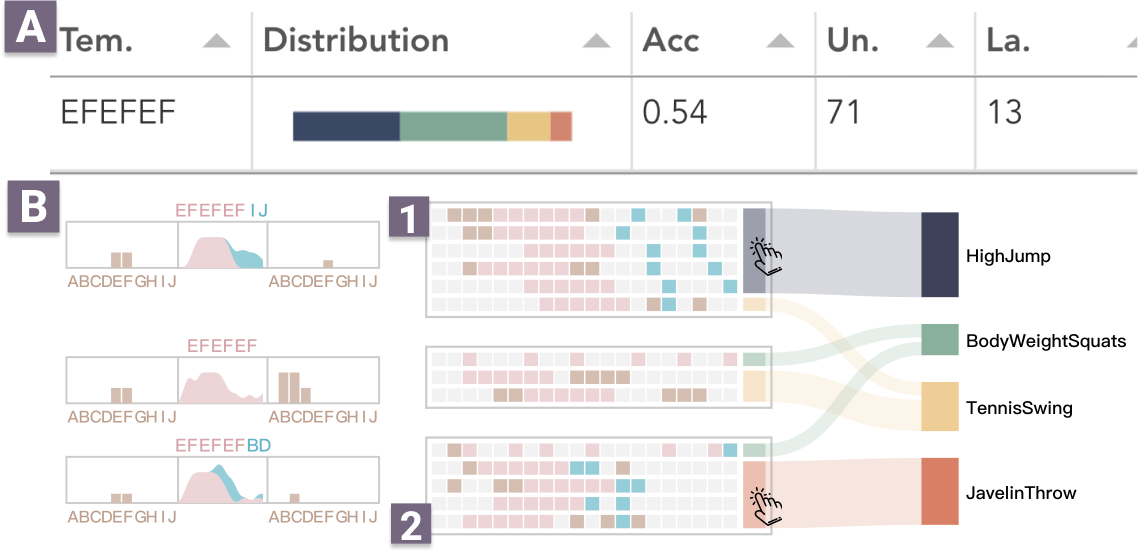}
    \caption{\rc{ (A) The template shared by \textit{High Jump} class and \textit{Javelin Throw} class. (B) The sub-templates in the \textit{Labeling View} with event temporal distribution of corresponding labeled videos. }}
    \label{fig: ucf_labelview}
\end{figure}

In the \textit{Labeling View}, three sub-templates were identified. 
By observing the flow width and color, \textbf{E6} noticed that the sub-template ``EFEFEFIJ'' (\cref{fig: ucf_labelview}B-1) predominantly contained videos from the \textit{High Jump} class, while the sub-template ``EFEFEFBD'' (\cref{fig: ucf_labelview}B-2) mainly included videos from the \textit{Javelin Throw} class (\textbf{R3}). 
This finding aligned with \textbf{E6}'s knowledge, as the event sequences following the approach stage captured the distinguishing characteristics of these two classes. The \textit{High Jump} class exhibited a body elevation event for takeoff, followed by a body depression event for landing. On the other hand, the \textit{Javelin Throw} action involved arm extension and adduction for the delivery and then release of the javelin.

\begin{figure}[ht]
    \centering
    \includegraphics[width=0.9\linewidth]{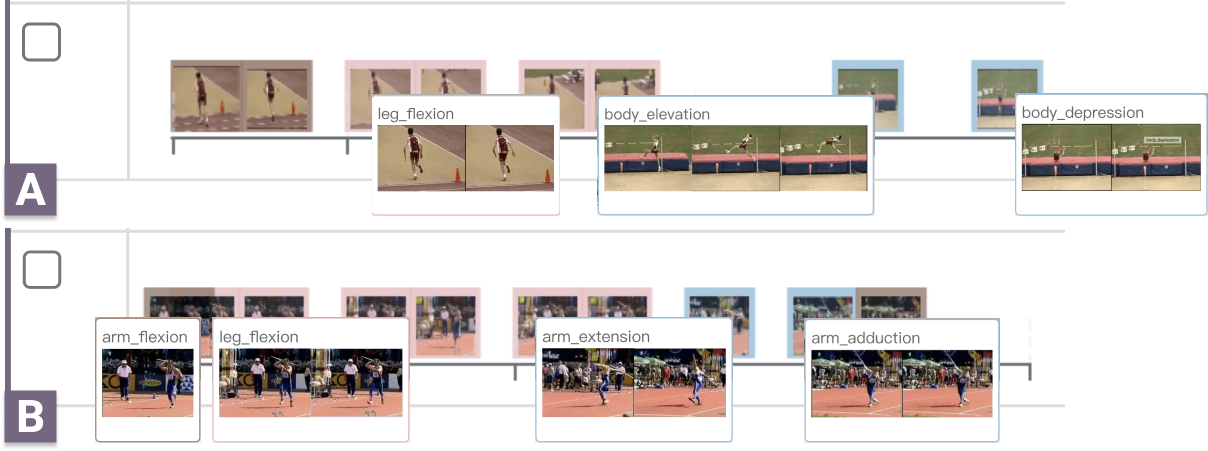}
    \caption{\rc{Two examples of retrieved videos for \textit{High Jump} class and \textit{Javelin Throw} class. (A) the video labeled as \textit{High Jump}. (B) the video labeled as \textit{Javelin Throw}.}}
    \label{fig:ucf_video}
\end{figure}

\textbf{E6} then clicked on the colored bars at the end of the two clusters respectively to retrieve similar unlabeled videos for labeling from the below video list. \textbf{E6} randomly checked several videos (\cref{fig:ucf_video}), observing the event sequence through the frame border color and hovering on the keyframes to unfold the frame sequences for quick video browsing (\textbf{R1}). After selecting the matched videos, he applied the labels to them at once (\textbf{R3}).
After supplementing more samples for the \textit{High Jump} class and \textit{Javelin Throw} class, \textbf{E6} initiated model retraining, resulting in noticeable improvements in performance for these two classes (\rc{+1.25\%} for the \textit{High Jump} class and \rc{+3.47\%} for the \textit{Javelin Throw} class). 

\textbf{E6} proceeded to program other classes with relatively low performance, such as\textit{Tennis Swing} and \textit{Lunges}. 
After 8 iterations with 304 videos being labeled, the model finally achieved an overall F1 score of 93.98\% on the test set (\textbf{R4}). 

\textbf{Post Analysis}
We followed a similar practice in case one to quantitatively evaluate the labeling efficiency. 
On this public dataset, the active learning-based labeling approach requires labeling 496 samples to achieve an overall F1 score of 93.92\%. 
For time cost estimation, we referred to Ma\etal~\cite{ma2020sf}, which reported an average of 45s for video-level action labels in a 60s video. 
Therefore, the time cost for the active learning-based approach is computed as 0.75 * (total time length of 496 labeled samples), which is 0.8h. 
In contrast, \textbf{E6} finished the whole programming in 0.5h. The results are listed in~\cref{tab: label_accuracy}, which further validates the efficiency of \name.
}

\subsection{Expert Interviews}
We further conducted semi-structured individual interviews with three ML practitioners (\textbf{P1-P3}) from the project development team, who have more than four years of experience in developing and operating ML models for video applications. 
\rc{While familiar with the project context, non of them had known or tried the system before the interview.}
The interviews began with an introduction to the research background and system designs.
Then we demonstrated system workflow and usage with specific examples~\cite{yange2023example}.
After the demonstration, we asked the practitioners to freely explore and try the system for programming on the real dataset, and express their thoughts, findings and suggestions in a think-aloud protocol.
\rc{We also collected feedback from \textbf{E6} during the second case study.}
The feedback collected was categorized into the following three perspectives: 

\textbf{System workflow} 
All participants confirmed the effectiveness of using human-understandable events to represent video data, which is ``\textit{intuitive and useful to understand video content}''.
They also appreciated the idea of extracting event sequential patterns as programming guide templates. 
\rc{\textbf{E6} commented, ``\textit{This tool is pretty helpful for labeling and analyzing sports tactics, as the event order directly determines the tactic type.}''}
Furthermore, the participants valued the tool's ability to efficiently search and retrieve videos from large-scale video datasets through flexible event composition and assembly. They emphasized that this is particularly ``\textit{important and needed}'' in real-world work scenarios, which allows them to retrieve video data samples at scale for model building and steering with minimal effort and cost.

\textbf{Visual designs and interactions} 
Overall, the practitioners reported the system is ``\textit{easy to use}'' with intuitive visual designs and smooth interactions. 
The \textit{Labeling View} is favored by all participants, where they can ``\textit{grasp the video content by glancing at the keyframes}''.
\rwf{\textbf{P2} appreciated the sorting of videos based on similarity, making it easy to identify the most matched videos and apply labels in batches conveniently.}
The design of \textit{Template View} is also well-received, especially for its rich interactions, enabling ``\textit{efficient template exploration based on different metrics.}''
\textbf{P3} expressed a liking for the projection design in the \textit{Info View}, with the intuitive background error heatmap and useful lasso interaction for selecting video groups of interest.
Nevertheless, the participants found the \textit{Labeling View} design somewhat complex, as it contained a lot of information, requiring some time for them to grasp. 

\textbf{Suggestions for improvement}
\textbf{P1} expressed the need to save the history of all selected templates for future reference. 
He also proposed that more events could be included to provide a more exhaustive summarization of the video content, while acknowledging the importance of focusing on critical ones. 
\textbf{P2} suggested that the system could provide real-time operation guidance and suggestions to reduce the learning curve.
He also mentioned that more advanced strategies are needed to resolve label conflicts, which are currently being handled manually.
\textbf{P3} recommended that the system should support adjusting more parameters such as learning rate and batch size on the interface.
\rc{\textbf{E6} suggested the use of semantic meaningful icons or abbreviations to enhance the intuitive understanding of events.}

\section{Discussion}
\label{sec:discussion}
During the development process of \name, we have gained insights, identified limitations, and got inspirations for future exploration.

\xingbo{
\textbf{Data-centric approach for video data programming with labeling templates}
Data programming adopts a data-centric perspective to enhance data quality at scale, enabling model steering under the supervision of users' domain knowledge.
While previous works have focused on temporal pattern labeling~\cite{lekschas2020peax} or static spatial relationships in images~\cite{hoque2022visual}, they fall short in handling the rich spatial and temporal semantic information present in videos.
Our work overcomes these limitations by utilizing semantic-rich events to compose labeling functions.
We employ compact labeling templates to summarize diverse events and their intricate temporal relationships, helping users to understand video data characteristics and identify semantic meaningful ones for labeling target data classes.
This ``video-event-template'' abstraction process effectively elicits users' high-level domain knowledge for data labeling and model training.
Currently, our templates mainly consider the semantics of event types and temporal orders. Future works can consider more complex semantics involving event characteristics like duration and object interactions. 
Meanwhile, when exploring different templates to distill meanings of event compositions, there often exists a trade-off between coverage and meaningfulness.
Some templates may cover a large number of instances but introduce some noisy and meaningless ones, requiring greater effort for validation.
On the contrary, some templates can accurately reflect the semantic meaning of a target data class but cover only a few instances. 
Future systems can also consider adaptive designs of templates that strike a balance between coverage and meaningfulness.
}

\textbf{System generalizability}
\rc{The proposed generic labeling workflow is capable of accommodating various tasks beyond classification with minor modifications.
For example, for temporal action localization (which seeks to identify the interval of a specific activity in untrimmed videos), \name can match the activity with its representative event sequences to provide a rough estimate of time spans.
Then, the \textit{Labeling View} can be revised to enable zooming in on the fine-grained components of the start and end events for precise start and end timestamp annotation.
For tasks such as video retrieval~\cite{yang2021induce} and generation~\cite{price2022unweavenet}, the template mining algorithm and the \textit{Template View} design can be directly employed to define and compose sequential event relationships flexibly.}
Moreover, considering the fundamental role of event sequences in video data, \name is transferable to a wide range of applications and video types.
For example, domains such as social science and behavior psychology share similar requirements for building models to analyze user behaviors and interactions in recorded experiment videos. The system also readily supports needs such as sports tactics analysis, tutorial video understanding, and surgical video comparison.
\rc{Furthermore, events can be compiled in different ways to create new classes flexibly for new use cases. For instance, altering the cooking order of ingredients can build templates for new recipes.}

\rc{\textbf{Event extraction effectiveness}
Defining atomic events properly relies on domain knowledge, task specifics, and suitable algorithms, considering the hierarchical nature of events.
For instance, a cooking video for ``preparing a salad'' involves atomic events such as chopping vegetables, tossing, and dressing the salad. 
These high-level events can either be detected by action recognition~\cite{2020mmaction2}, or further decomposed as hand movements and object manipulations that can be deduced through heuristics~\cite{ji2020action}. 
Visual-language models have also emerged as a powerful tool for tagging semantic concepts~\cite{wang2021actionclip, feng2023promptmagician} (\eg objects, actions, and scenes), offering new possibilities for capturing complex higher-level semantic events.
Apart from employing more powerful models, visualizations should be designed for summarizing high-level event semantics and facilitating intuitive reviews of detailed video frames.
High-level semantics representations (\eg object-scene graphs~\cite{ou2022object}) can also benefit from novel visual designs to analyze event relationships.
\rwf{Considering the impact of imperfect algorithms, camera movements, and view occlusions on event detection, incorporating more robust algorithms and uncertainty visualization techniques can enhance the system's resilience and reliability.}
}

\rc{\textbf{System scalability}
In terms of visual design, when the number of classes and event categories reaches tens and hundreds, 
it will lead to a long template list and visual clutters in the distribution bar charts in the \textit{Template View}. 
To address this, \name offers sorting and filtering options based on multiple thresholds and metrics, allowing users to quickly explore and locate templates of interest.
Similarly, the Sankey diagram design in the \textit{Labeling View} may become visually cluttered with a large number of classes. However, since experts often need to compare only a few classes at the same time, the \textit{Labeling View} can satisfy their requirements.
In the future, we plan to implement multi-level grouping strategies, together with hierarchical visualization and interaction techniques to further enhance visual scalability.
For example, we can group classes and events based on some taxonomies, themes, or model performance. Then users can explore and program different video data subsets that contain a few categories of interest.}
\textbf{Limitations and future works}
\rc{Currently, \name is designed for discrete events and could face challenges with datasets featuring longer, overlapping events. In such cases, events could be weighted based on their significance or aggregated into compounds (\eg $A(AB)B \rightarrow ACB$) to retain sequential patterns. However, these adjustments may complicate template mining, making additional research necessary for overlapping events~\cite{related_event_zeng_2020}.}
Additionally, the system currently supports video programming through visual channels only, while some video analyses could benefit from incorporating concurrent audio and speech information~\cite{related_event_wong_2023}. 
Therefore, efficient methods of encoding and complementing multimodal information~\cite{related_event_mm_wang_2022} for programming are worth exploring.
Furthermore, the current system is designed for single-person operations. To enable collaborative programming, it is important to explore methods for efficiently resolving label conflict and maintaining consistent labeling quality in future work.
\section{Conclusion}
\rwf{This paper presents \name, a novel visual analytics approach that extracts and externalizes video event composition knowledge to streamline video data programming. 
The conducted two case studies and expert interviews validate the system's efficiency and effectiveness for video data supplementation and model steering.
Meanwhile, the development and evaluation of \name reveal several promising future research directions, including integrating more complex event attributes, balancing template coverage and meaningfulness, and exploring multimodal and collaborative video programming techniques.}
\acknowledgments{%
The authors would like to thank the anonymous reviewers for their constructive and insightful comments. This work is partially supported by Hong Kong ITF grant PRP/001/21FX.
	
}

\bibliographystyle{src/abbrv-doi-hyperref}

\bibliography{main}

\begin{thebibliography}{10}

\bibitem{abernethy2013biophysical}
B.~Abernethy, V.~Kippers, and S.~J. Hanrahan.
\newblock {\em Biophysical foundations of human movement}.
\newblock Human Kinetics, 2013.

\bibitem{related_event_afzal_2023}
S.~Afzal, S.~Ghani, M.~M. Hittawe, S.~F. Rashid, O.~M. Knio, M.~Hadwiger, and
  I.~Hoteit.
\newblock Visualization and visual analytics approaches for image and video
  datasets: A survey.
\newblock {\em ACM Transactions on Interactive Intelligent Systems},
  13(1):1--41, 2023. \href{https://doi.org/10.1145/3576935}
{doi: {{%
10\hspace{.1pt}\discretionary{.}{%
}{.}\hspace{.4pt}1145\discretionary{/}{%
}{/}3576935}}}


\bibitem{bagher-2018-multimodal}
A.~Bagher~Zadeh, P.~P. Liang, S.~Poria, E.~Cambria, and L.-P. Morency.
\newblock Multimodal language analysis in the wild: {CMU}-{MOSEI} dataset and
  interpretable dynamic fusion graph.
\newblock In {\em Proc. ACL}, pp. 2236--2246. ACL, Melbourne, Australia, 2018.
  \href{https://doi.org/10.18653/v1/P18-1208}
{doi: {{%
10\hspace{.1pt}\discretionary{.}{%
}{.}\hspace{.4pt}18653\discretionary{/}{%
}{/}v1\discretionary{/}{%
}{/}P18\discretionary{%
}{-}{-}1208}}}


\bibitem{baltruvsaitis2016openface}
T.~Baltrušaitis, P.~Robinson, and L.-P. Morency.
\newblock Openface: An open source facial behavior analysis toolkit.
\newblock In {\em Proc. WACV}, pp. 1--10. IEEE, Los Alamitos, 2016.
  \href{https://doi.org/10.1109/WACV.2016.7477553}
{doi: {{%
10\hspace{.1pt}\discretionary{.}{%
}{.}\hspace{.4pt}1109\discretionary{/}{%
}{/}WACV\hspace{.1pt}\discretionary{.}{%
}{.}\hspace{.4pt}2016\hspace{.1pt}\discretionary{.}{%
}{.}\hspace{.4pt}7477553}}}


\bibitem{bernard2021taxonomy}
J.~Bernard, M.~Hutter, M.~Sedlmair, M.~Zeppelzauer, and T.~Munzner.
\newblock A taxonomy of property measures to unify active learning and
  human-centered approaches to data labeling.
\newblock {\em ACM Transactions on Interactive Intelligent Systems},
  11(3–4):1--42, sep 2021. \href{https://doi.org/10.1145/3439333}
{doi: {{%
10\hspace{.1pt}\discretionary{.}{%
}{.}\hspace{.4pt}1145\discretionary{/}{%
}{/}3439333}}}


\bibitem{bernard2018towards}
J.~Bernard, M.~Zeppelzauer, M.~Lehmann, M.~Müller, and M.~Sedlmair.
\newblock Towards user-centered active learning algorithms.
\newblock {\em Computer Graphics Forum}, 37(3):121--132, 2018.
  \href{https://doi.org/10.1111/cgf.13406}
{doi: {{%
10\hspace{.1pt}\discretionary{.}{%
}{.}\hspace{.4pt}1111\discretionary{/}{%
}{/}cgf\hspace{.1pt}\discretionary{.}{%
}{.}\hspace{.4pt}13406}}}


\bibitem{related_event_blascheck_2016}
T.~Blascheck, F.~Beck, S.~Baltes, T.~Ertl, and D.~Weiskopf.
\newblock Visual analysis and coding of data-rich user behavior.
\newblock In {\em Proc. VAST}, pp. 141--150. IEEE, Los Alamitos, 2016.
  \href{https://doi.org/10.1109/VAST.2016.7883520}
{doi: {{%
10\hspace{.1pt}\discretionary{.}{%
}{.}\hspace{.4pt}1109\discretionary{/}{%
}{/}VAST\hspace{.1pt}\discretionary{.}{%
}{.}\hspace{.4pt}2016\hspace{.1pt}\discretionary{.}{%
}{.}\hspace{.4pt}7883520}}}


\bibitem{hand_detection}
Cansik.
\newblock "yolo-hand-detection," github.com.
\newblock \url{https://github.com/cansik/yolo-hand-detection}, 2022.
\newblock (accessed Jun. 21, 2023).

\bibitem{cao2019openpose}
Z.~{Cao}, G.~{Hidalgo Martinez}, T.~{Simon}, S.~{Wei}, and Y.~A. {Sheikh}.
\newblock Openpose: Realtime multi-person 2d pose estimation using part
  affinity fields.
\newblock {\em IEEE Transactions on Pattern Analysis and Machine Intelligence},
  43(1):172--186, 2019. \href{https://doi.org/10.1109/TPAMI.2019.2929257}
{doi: {{%
10\hspace{.1pt}\discretionary{.}{%
}{.}\hspace{.4pt}1109\discretionary{/}{%
}{/}TPAMI\hspace{.1pt}\discretionary{.}{%
}{.}\hspace{.4pt}2019\hspace{.1pt}\discretionary{.}{%
}{.}\hspace{.4pt}2929257}}}


\bibitem{carreira2017quo}
J.~Carreira and A.~Zisserman.
\newblock Quo vadis, action recognition? a new model and the kinetics dataset.
\newblock In {\em Proc. CVPR}, pp. 4724--4733. IEEE, Los Alamitos, 2017.
  \href{https://doi.org/10.1109/CVPR.2017.502}
{doi: {{%
10\hspace{.1pt}\discretionary{.}{%
}{.}\hspace{.4pt}1109\discretionary{/}{%
}{/}CVPR\hspace{.1pt}\discretionary{.}{%
}{.}\hspace{.4pt}2017\hspace{.1pt}\discretionary{.}{%
}{.}\hspace{.4pt}502}}}


\bibitem{chen2021interactive}
C.~Chen, Z.~Wang, J.~Wu, X.~Wang, L.-Z. Guo, Y.-F. Li, and S.~Liu.
\newblock Interactive graph construction for graph-based semi-supervised
  learning.
\newblock {\em IEEE Transactions on Visualization and Computer Graphics},
  27(9):3701--3716, 2021. \href{https://doi.org/10.1109/TVCG.2021.3084694}
{doi: {{%
10\hspace{.1pt}\discretionary{.}{%
}{.}\hspace{.4pt}1109\discretionary{/}{%
}{/}TVCG\hspace{.1pt}\discretionary{.}{%
}{.}\hspace{.4pt}2021\hspace{.1pt}\discretionary{.}{%
}{.}\hspace{.4pt}3084694}}}


\bibitem{chen2022towards}
C.~Chen, J.~Wu, X.~Wang, S.~Xiang, S.-H. Zhang, Q.~Tang, and S.~Liu.
\newblock Towards better caption supervision for object detection.
\newblock {\em IEEE Transactions on Visualization and Computer Graphics},
  28(4):1941--1954, 2022. \href{https://doi.org/10.1109/TVCG.2021.3138933}
{doi: {{%
10\hspace{.1pt}\discretionary{.}{%
}{.}\hspace{.4pt}1109\discretionary{/}{%
}{/}TVCG\hspace{.1pt}\discretionary{.}{%
}{.}\hspace{.4pt}2021\hspace{.1pt}\discretionary{.}{%
}{.}\hspace{.4pt}3138933}}}


\bibitem{chen2020oodanalyzer}
C.~Chen, J.~Yuan, Y.~Lu, Y.~Liu, H.~Su, S.~Yuan, and S.~Liu.
\newblock {OoDAnalyzer}: Interactive analysis of out-of-distribution samples.
\newblock {\em IEEE Transactions on Visualization and Computer Graphics},
  27(7):3335--3349, 2021. \href{https://doi.org/10.1109/TVCG.2020.2973258}
{doi: {{%
10\hspace{.1pt}\discretionary{.}{%
}{.}\hspace{.4pt}1109\discretionary{/}{%
}{/}TVCG\hspace{.1pt}\discretionary{.}{%
}{.}\hspace{.4pt}2020\hspace{.1pt}\discretionary{.}{%
}{.}\hspace{.4pt}2973258}}}


\bibitem{chen2018sequence}
Y.~Chen, P.~Xu, and L.~Ren.
\newblock Sequence synopsis: Optimize visual summary of temporal event data.
\newblock {\em IEEE Transactions on Visualization and Computer Graphics},
  24(1):45--55, 2018. \href{https://doi.org/10.1109/TVCG.2017.2745083}
{doi: {{%
10\hspace{.1pt}\discretionary{.}{%
}{.}\hspace{.4pt}1109\discretionary{/}{%
}{/}TVCG\hspace{.1pt}\discretionary{.}{%
}{.}\hspace{.4pt}2017\hspace{.1pt}\discretionary{.}{%
}{.}\hspace{.4pt}2745083}}}


\bibitem{related_event_chen_2022}
Z.~Chen et~al.
\newblock Augmenting sports videos with viscommentator.
\newblock {\em IEEE Transactions on Visualization and Computer Graphics},
  28(1):824--834, 2022. \href{https://doi.org/10.1109/TVCG.2021.3114806}
{doi: {{%
10\hspace{.1pt}\discretionary{.}{%
}{.}\hspace{.4pt}1109\discretionary{/}{%
}{/}TVCG\hspace{.1pt}\discretionary{.}{%
}{.}\hspace{.4pt}2021\hspace{.1pt}\discretionary{.}{%
}{.}\hspace{.4pt}3114806}}}


\bibitem{choi2019aila}
M.~Choi, C.~Park, S.~Yang, Y.~Kim, J.~Choo, and S.~R. Hong.
\newblock Aila: Attentive interactive labeling assistant for document
  classification through attention-based deep neural networks.
\newblock In {\em Proc. CHI}, pp. 1--12. ACM, New York, 2019.
  \href{https://doi.org/10.1145/3290605.3300460}
{doi: {{%
10\hspace{.1pt}\discretionary{.}{%
}{.}\hspace{.4pt}1145\discretionary{/}{%
}{/}3290605\hspace{.1pt}\discretionary{.}{%
}{.}\hspace{.4pt}3300460}}}


\bibitem{2020mmaction2}
M.~Contributors.
\newblock Openmmlab's next generation video understanding toolbox and
  benchmark.
\newblock \url{https://github.com/open-mmlab/mmaction2}, 2020.

\bibitem{related_event_dasiopoulou_2011}
S.~Dasiopoulou, E.~Giannakidou, G.~Litos, P.~Malasioti, and Y.~Kompatsiaris.
\newblock {\em A Survey of Semantic Image and Video Annotation Tools}, pp.
  196--239.
\newblock Springer, Cham, Switzerland, 2011.
  \href{https://doi.org/10.1007/978-3-642-20795-2_8}
{doi: {{%
10\hspace{.1pt}\discretionary{.}{%
}{.}\hspace{.4pt}1007\discretionary{/}{%
}{/}978\discretionary{%
}{-}{-}3\discretionary{%
}{-}{-}642\discretionary{%
}{-}{-}20795\discretionary{%
}{-}{-}2\_8}}}


\bibitem{related_event_deng_2021}
D.~Deng et~al.
\newblock Eventanchor: Reducing human interactions in event annotation of
  racket sports videos.
\newblock In {\em Proc. CHI}, pp. 1--13. ACM, New York, 2021.
  \href{https://doi.org/10.1145/3411764.3445431}
{doi: {{%
10\hspace{.1pt}\discretionary{.}{%
}{.}\hspace{.4pt}1145\discretionary{/}{%
}{/}3411764\hspace{.1pt}\discretionary{.}{%
}{.}\hspace{.4pt}3445431}}}


\bibitem{jiao2022new}
L.~J. et~al.
\newblock New generation deep learning for video object detection: A survey.
\newblock {\em IEEE Transactions on Neural Networks and Learning Systems},
  33(8):3195--3215, 2022. \href{https://doi.org/10.1109/TNNLS.2021.3053249}
{doi: {{%
10\hspace{.1pt}\discretionary{.}{%
}{.}\hspace{.4pt}1109\discretionary{/}{%
}{/}TNNLS\hspace{.1pt}\discretionary{.}{%
}{.}\hspace{.4pt}2021\hspace{.1pt}\discretionary{.}{%
}{.}\hspace{.4pt}3053249}}}


\bibitem{feng2023promptmagician}
Y.~Feng, X.~Wang, K.~K. Wong, S.~Wang, Y.~Lu, M.~Zhu, B.~Wang, and W.~Chen.
\newblock Promptmagician: Interactive prompt engineering for text-to-image
  creation.
\newblock {\em arXiv}, 2023. \href{https://doi.org/10.48550/arXiv.2307.09036}
{doi: {{%
10\hspace{.1pt}\discretionary{.}{%
}{.}\hspace{.4pt}48550\discretionary{/}{%
}{/}arXiv\hspace{.1pt}\discretionary{.}{%
}{.}\hspace{.4pt}2307\hspace{.1pt}\discretionary{.}{%
}{.}\hspace{.4pt}09036}}}


\bibitem{grimmeisen2022visgil}
B.~Grimmeisen, M.~Chegini, and A.~Theissler.
\newblock Visgil: machine learning-based visual guidance for interactive
  labeling.
\newblock {\em The Visual Computer}, pp. 1--23, 2022.
  \href{https://doi.org/10.1007/s00371-022-02648-2}
{doi: {{%
10\hspace{.1pt}\discretionary{.}{%
}{.}\hspace{.4pt}1007\discretionary{/}{%
}{/}s00371\discretionary{%
}{-}{-}022\discretionary{%
}{-}{-}02648\discretionary{%
}{-}{-}2}}}


\bibitem{grunwald2007mdl}
P.~D. Gr\"{u}nwald.
\newblock {\em The Minimum Description Length Principle (Adaptive Computation
  and Machine Learning)}.
\newblock The MIT Press, 2007.

\bibitem{related_event_halter_2019}
G.~Halter, R.~Ballester-Ripoll, B.~Flueckiger, and R.~Pajarola.
\newblock Vian: A visual annotation tool for film analysis.
\newblock {\em Computer Graphics Forum}, 38(3):119--129, 2019.
  \href{https://doi.org/10.1111/cgf.13676}
{doi: {{%
10\hspace{.1pt}\discretionary{.}{%
}{.}\hspace{.4pt}1111\discretionary{/}{%
}{/}cgf\hspace{.1pt}\discretionary{.}{%
}{.}\hspace{.4pt}13676}}}


\bibitem{hamill2006biomechanical}
J.~Hamill and K.~M. Knutzen.
\newblock {\em Biomechanical basis of human movement}.
\newblock Lippincott Williams \& Wilkins, 2006.

\bibitem{hoque2022visual}
M.~N. Hoque, W.~He, A.~K. Shekar, L.~Gou, and L.~Ren.
\newblock Visual concept programming: A visual analytics approach to injecting
  human intelligence at scale.
\newblock {\em IEEE Transactions on Visualization and Computer Graphics},
  29(1):74--83, 2023. \href{https://doi.org/10.1109/TVCG.2022.3209466}
{doi: {{%
10\hspace{.1pt}\discretionary{.}{%
}{.}\hspace{.4pt}1109\discretionary{/}{%
}{/}TVCG\hspace{.1pt}\discretionary{.}{%
}{.}\hspace{.4pt}2022\hspace{.1pt}\discretionary{.}{%
}{.}\hspace{.4pt}3209466}}}


\bibitem{hosseininasab2019constraint}
A.~Hosseininasab, W.-J. van Hoeve, and A.~A. Cire.
\newblock Constraint-based sequential pattern mining with decision diagrams.
\newblock In {\em Proc. AAAI}, pp. 1495--1502. AAAI Press, 2019.
  \href{https://doi.org/10.1609/aaai.v33i01.33011495}
{doi: {{%
10\hspace{.1pt}\discretionary{.}{%
}{.}\hspace{.4pt}1609\discretionary{/}{%
}{/}aaai\hspace{.1pt}\discretionary{.}{%
}{.}\hspace{.4pt}v33i01\hspace{.1pt}\discretionary{.}{%
}{.}\hspace{.4pt}33011495}}}


\bibitem{related_event_hoferlin_2015}
B.~Höferlin, M.~Höferlin, G.~Heidemann, and D.~Weiskopf.
\newblock Scalable video visual analytics.
\newblock {\em Information Visualization}, 14(1):10--26, 2015.
  \href{https://doi.org/10.1177/1473871613488571}
{doi: {{%
10\hspace{.1pt}\discretionary{.}{%
}{.}\hspace{.4pt}1177\discretionary{/}{%
}{/}1473871613488571}}}


\bibitem{related_event_hoferlin_2012}
B.~Höferlin, R.~Netzel, M.~Höferlin, D.~Weiskopf, and G.~Heidemann.
\newblock Inter-active learning of ad-hoc classifiers for video visual
  analytics.
\newblock In {\em Proc. VAST}, pp. 23--32. IEEE, Los Alamitos, 2012.
  \href{https://doi.org/10.1109/VAST.2012.6400492}
{doi: {{%
10\hspace{.1pt}\discretionary{.}{%
}{.}\hspace{.4pt}1109\discretionary{/}{%
}{/}VAST\hspace{.1pt}\discretionary{.}{%
}{.}\hspace{.4pt}2012\hspace{.1pt}\discretionary{.}{%
}{.}\hspace{.4pt}6400492}}}


\bibitem{ji2020action}
J.~Ji, R.~Krishna, F.-F. Li, and J.~C. Niebles.
\newblock Action genome: Actions as compositions of spatio-temporal scene
  graphs.
\newblock In {\em Proc. CVPR}, pp. 10236--10247. IEEE, Los Alamitos, 2020.
  \href{https://doi.org/10.1109/CVPR42600.2020.01025}
{doi: {{%
10\hspace{.1pt}\discretionary{.}{%
}{.}\hspace{.4pt}1109\discretionary{/}{%
}{/}CVPR42600\hspace{.1pt}\discretionary{.}{%
}{.}\hspace{.4pt}2020\hspace{.1pt}\discretionary{.}{%
}{.}\hspace{.4pt}01025}}}


\bibitem{jia2021towards}
S.~Jia, Z.~Li, N.~Chen, and J.~Zhang.
\newblock Towards visual explainable active learning for zero-shot
  classification.
\newblock {\em IEEE Transactions on Visualization and Computer Graphics},
  28(1):791--801, 2022. \href{https://doi.org/10.1109/TVCG.2021.3114793}
{doi: {{%
10\hspace{.1pt}\discretionary{.}{%
}{.}\hspace{.4pt}1109\discretionary{/}{%
}{/}TVCG\hspace{.1pt}\discretionary{.}{%
}{.}\hspace{.4pt}2021\hspace{.1pt}\discretionary{.}{%
}{.}\hspace{.4pt}3114793}}}


\bibitem{khayat2019vassl}
M.~Khayat, M.~Karimzadeh, J.~Zhao, and D.~S. Ebert.
\newblock Vassl: A visual analytics toolkit for social spambot labeling.
\newblock {\em IEEE Transactions on Visualization and Computer Graphics},
  26(1):874--883, 2020. \href{https://doi.org/10.1109/TVCG.2019.2934266}
{doi: {{%
10\hspace{.1pt}\discretionary{.}{%
}{.}\hspace{.4pt}1109\discretionary{/}{%
}{/}TVCG\hspace{.1pt}\discretionary{.}{%
}{.}\hspace{.4pt}2019\hspace{.1pt}\discretionary{.}{%
}{.}\hspace{.4pt}2934266}}}


\bibitem{kurby2008segmentation}
C.~A. Kurby and J.~M. Zacks.
\newblock Segmentation in the perception and memory of events.
\newblock {\em Trends in cognitive sciences}, 12(2):72--79, 2008.
  \href{https://doi.org/10.1016/j.tics.2007.11.004}
{doi: {{%
10\hspace{.1pt}\discretionary{.}{%
}{.}\hspace{.4pt}1016\discretionary{/}{%
}{/}j\hspace{.1pt}\discretionary{.}{%
}{.}\hspace{.4pt}tics\hspace{.1pt}\discretionary{.}{%
}{.}\hspace{.4pt}2007\hspace{.1pt}\discretionary{.}{%
}{.}\hspace{.4pt}11\hspace{.1pt}\discretionary{.}{%
}{.}\hspace{.4pt}004}}}


\bibitem{kurzhals2016visual}
K.~Kurzhals, M.~Hlawatsch, C.~Seeger, and D.~Weiskopf.
\newblock Visual analytics for mobile eye tracking.
\newblock {\em IEEE Transactions on Visualization and Computer Graphics},
  23(1):301--310, 2017. \href{https://doi.org/10.1109/TVCG.2016.2598695}
{doi: {{%
10\hspace{.1pt}\discretionary{.}{%
}{.}\hspace{.4pt}1109\discretionary{/}{%
}{/}TVCG\hspace{.1pt}\discretionary{.}{%
}{.}\hspace{.4pt}2016\hspace{.1pt}\discretionary{.}{%
}{.}\hspace{.4pt}2598695}}}


\bibitem{lasecki2014glance}
W.~S. Lasecki, M.~Gordon, D.~Koutra, M.~F. Jung, S.~P. Dow, and J.~P. Bigham.
\newblock Glance: Rapidly coding behavioral video with the crowd.
\newblock In {\em Proc. UIST}, pp. 551--562. ACM, New York, 2014.
  \href{https://doi.org/10.1145/2642918.2647367}
{doi: {{%
10\hspace{.1pt}\discretionary{.}{%
}{.}\hspace{.4pt}1145\discretionary{/}{%
}{/}2642918\hspace{.1pt}\discretionary{.}{%
}{.}\hspace{.4pt}2647367}}}


\bibitem{lekschas2020peax}
F.~Lekschas, B.~Peterson, D.~Haehn, E.~Ma, N.~Gehlenborg, and H.~Pfister.
\newblock Peax: Interactive visual pattern search in sequential data using
  unsupervised deep representation learning.
\newblock {\em Computer Graphics Forum}, 39(3):167--179, 2020.
  \href{https://doi.org/10.1111/cgf.13971}
{doi: {{%
10\hspace{.1pt}\discretionary{.}{%
}{.}\hspace{.4pt}1111\discretionary{/}{%
}{/}cgf\hspace{.1pt}\discretionary{.}{%
}{.}\hspace{.4pt}13971}}}


\bibitem{related_event_li_2021}
H.~Li, M.~Xu, Y.~Wang, H.~Wei, and H.~Qu.
\newblock A visual analytics approach to facilitate the proctoring of online
  exams.
\newblock In {\em Proc. CHI}, pp. 1--17. ACM, New York, 2021.
  \href{https://doi.org/10.1145/3411764.3445294}
{doi: {{%
10\hspace{.1pt}\discretionary{.}{%
}{.}\hspace{.4pt}1145\discretionary{/}{%
}{/}3411764\hspace{.1pt}\discretionary{.}{%
}{.}\hspace{.4pt}3445294}}}


\bibitem{li2021weakly}
J.~Li, H.~Ding, J.~Shang, J.~McAuley, and Z.~Feng.
\newblock Weakly supervised named entity tagging with learnable logical rules.
\newblock In {\em Proc. IJCNLP}, pp. 4568--4581. ACL, New York, 2021.
  \href{https://doi.org/10.18653/v1/2021.acl-long.352}
{doi: {{%
10\hspace{.1pt}\discretionary{.}{%
}{.}\hspace{.4pt}18653\discretionary{/}{%
}{/}v1\discretionary{/}{%
}{/}2021\hspace{.1pt}\discretionary{.}{%
}{.}\hspace{.4pt}acl\discretionary{%
}{-}{-}long\hspace{.1pt}\discretionary{.}{%
}{.}\hspace{.4pt}352}}}


\bibitem{li2022uniformer}
K.~Li, Y.~Wang, P.~Gao, G.~Song, Y.~Liu, H.~Li, and Y.~Qiao.
\newblock Uniformer: Unified transformer for efficient spatiotemporal
  representation learning.
\newblock {\em arXiv}, 2022. \href{https://doi.org/10.48550/arXiv.2201.04676}
{doi: {{%
10\hspace{.1pt}\discretionary{.}{%
}{.}\hspace{.4pt}48550\discretionary{/}{%
}{/}arXiv\hspace{.1pt}\discretionary{.}{%
}{.}\hspace{.4pt}2201\hspace{.1pt}\discretionary{.}{%
}{.}\hspace{.4pt}04676}}}


\bibitem{liang2023multiviz}
P.~P. Liang, Y.~Lyu, G.~Chhablani, N.~Jain, Z.~Deng, X.~Wang, L.-P. Morency,
  and R.~Salakhutdinov.
\newblock Multiviz: Towards visualizing and understanding multimodal models.
\newblock In {\em International Conference on Learning Representations}, 2023.

\bibitem{lin2014microsoft}
T.-Y. Lin et~al.
\newblock Microsoft coco: Common objects in context.
\newblock In {\em Proc. ECCV}, pp. 740--755. Springer, Cham, Switzerland, 2014.
  \href{https://doi.org/10.1007/978-3-319-10602-1_48}
{doi: {{%
10\hspace{.1pt}\discretionary{.}{%
}{.}\hspace{.4pt}1007\discretionary{/}{%
}{/}978\discretionary{%
}{-}{-}3\discretionary{%
}{-}{-}319\discretionary{%
}{-}{-}10602\discretionary{%
}{-}{-}1\_48}}}


\bibitem{liu2018crowsourcing}
S.~Liu, C.~Chen, Y.~Lu, F.~Ouyang, and B.~Wang.
\newblock An interactive method to improve crowdsourced annotations.
\newblock {\em IEEE Transactions on Visualization and Computer Graphics},
  25(1):235--245, 2019. \href{https://doi.org/10.1109/tvcg.2018.2864843}
{doi: {{%
10\hspace{.1pt}\discretionary{.}{%
}{.}\hspace{.4pt}1109\discretionary{/}{%
}{/}tvcg\hspace{.1pt}\discretionary{.}{%
}{.}\hspace{.4pt}2018\hspace{.1pt}\discretionary{.}{%
}{.}\hspace{.4pt}2864843}}}


\bibitem{ma2020sf}
F.~Ma et~al.
\newblock Sf-net: Single-frame supervision for temporal action localization.
\newblock In {\em Proc. ECCV}, pp. 420--437. Springer, Cham, Switzerland, 2020.
  \href{https://doi.org/10.1007/978-3-030-58548-8_25}
{doi: {{%
10\hspace{.1pt}\discretionary{.}{%
}{.}\hspace{.4pt}1007\discretionary{/}{%
}{/}978\discretionary{%
}{-}{-}3\discretionary{%
}{-}{-}030\discretionary{%
}{-}{-}58548\discretionary{%
}{-}{-}8\_25}}}


\bibitem{related_event_maher_2022}
K.~Maher et~al.
\newblock E-ffective: A visual analytic system for exploring the emotion and
  effectiveness of inspirational speeches.
\newblock {\em IEEE Transactions on Visualization and Computer Graphics},
  28(1):508--517, 2022. \href{https://doi.org/10.1109/TVCG.2021.3114789}
{doi: {{%
10\hspace{.1pt}\discretionary{.}{%
}{.}\hspace{.4pt}1109\discretionary{/}{%
}{/}TVCG\hspace{.1pt}\discretionary{.}{%
}{.}\hspace{.4pt}2021\hspace{.1pt}\discretionary{.}{%
}{.}\hspace{.4pt}3114789}}}


\bibitem{mcinnes2020umap}
L.~McInnes, J.~Healy, and J.~Melville.
\newblock Umap: Uniform manifold approximation and projection for dimension
  reduction.
\newblock {\em arXiv}, 2020. \href{https://doi.org/10.48550/arXiv.1802.03426}
{doi: {{%
10\hspace{.1pt}\discretionary{.}{%
}{.}\hspace{.4pt}48550\discretionary{/}{%
}{/}arXiv\hspace{.1pt}\discretionary{.}{%
}{.}\hspace{.4pt}1802\hspace{.1pt}\discretionary{.}{%
}{.}\hspace{.4pt}03426}}}


\bibitem{smile_detection}
Meng1994412.
\newblock "smile\_detection," github.com.
\newblock \url{https://github.com/meng1994412/Smile_Detection}, 2022.
\newblock (accessed Jun. 21, 2023).

\bibitem{moehrmann2011improving}
J.~Moehrmann, S.~Bernstein, T.~Schlegel, G.~Werner, and G.~Heidemann.
\newblock Improving the usability of hierarchical representations for
  interactively labeling large image data sets.
\newblock In {\em Proc. HCII}, pp. 618--627. Springer, Cham, Switzerland, 2011.

\bibitem{periphery_plots}
B.~Morrow, T.~Manz, A.~E. Chung, N.~Gehlenborg, and D.~Gotz.
\newblock Periphery plots for contextualizing heterogeneous time-based charts.
\newblock In {\em IEEE Visualization Conference (VIS)}, pp. 1--5, 2019.
  \href{https://doi.org/10.1109/VISUAL.2019.8933582}
{doi: {{%
10\hspace{.1pt}\discretionary{.}{%
}{.}\hspace{.4pt}1109\discretionary{/}{%
}{/}VISUAL\hspace{.1pt}\discretionary{.}{%
}{.}\hspace{.4pt}2019\hspace{.1pt}\discretionary{.}{%
}{.}\hspace{.4pt}8933582}}}


\bibitem{ou2022object}
Y.~Ou, L.~Mi, and Z.~Chen.
\newblock Object-relation reasoning graph for action recognition.
\newblock In {\em Proc. CVPR}, pp. 20133--20142, 2022.

\bibitem{related_event_parry_2011}
M.~Parry, P.~Legg, D.~H. Chung, I.~Griffiths, and M.~Chen.
\newblock Hierarchical event selection for video storyboards with a case study
  on snooker video visualization.
\newblock {\em IEEE Transactions on Visualization and Computer Graphics},
  17(12):1747--1756, 2011. \href{https://doi.org/10.1109/TVCG.2011.208}
{doi: {{%
10\hspace{.1pt}\discretionary{.}{%
}{.}\hspace{.4pt}1109\discretionary{/}{%
}{/}TVCG\hspace{.1pt}\discretionary{.}{%
}{.}\hspace{.4pt}2011\hspace{.1pt}\discretionary{.}{%
}{.}\hspace{.4pt}208}}}


\bibitem{price2022unweavenet}
W.~Price, C.~Vondrick, and D.~Damen.
\newblock Unweavenet: Unweaving activity stories.
\newblock In {\em Proc. CVPR}, pp. 13770--13779. IEEE, Los Alamitos, 2022.
  \href{https://doi.org/10.1109/CVPR52688.2022.01340}
{doi: {{%
10\hspace{.1pt}\discretionary{.}{%
}{.}\hspace{.4pt}1109\discretionary{/}{%
}{/}CVPR52688\hspace{.1pt}\discretionary{.}{%
}{.}\hspace{.4pt}2022\hspace{.1pt}\discretionary{.}{%
}{.}\hspace{.4pt}01340}}}


\bibitem{ratner2017snorkel}
A.~Ratner, S.~H. Bach, H.~Ehrenberg, J.~Fries, S.~Wu, and C.~R\'{e}.
\newblock Snorkel: Rapid training data creation with weak supervision.
\newblock {\em Proc. VLDB Endow.}, pp. 269--282, 2017.
  \href{https://doi.org/10.14778/3157794.3157797}
{doi: {{%
10\hspace{.1pt}\discretionary{.}{%
}{.}\hspace{.4pt}14778\discretionary{/}{%
}{/}3157794\hspace{.1pt}\discretionary{.}{%
}{.}\hspace{.4pt}3157797}}}


\bibitem{ratner2016data}
A.~Ratner, C.~D. Sa, S.~Wu, D.~Selsam, and C.~R\'{e}.
\newblock Data programming: Creating large training sets, quickly.
\newblock In {\em Proc. NeurIPS}, pp. 3574--3582. Curran Associates Inc., Red
  Hook, NY, USA, 2016.

\bibitem{rooij2010mediatable}
O.~Rooij, J.~van Wijk, and M.~Worring.
\newblock Mediatable: Interactive categorization of multimedia collections.
\newblock {\em IEEE Computer Graphics and Applications}, 30(5):42--51, 2010.
  \href{https://doi.org/10.1109/MCG.2010.66}
{doi: {{%
10\hspace{.1pt}\discretionary{.}{%
}{.}\hspace{.4pt}1109\discretionary{/}{%
}{/}MCG\hspace{.1pt}\discretionary{.}{%
}{.}\hspace{.4pt}2010\hspace{.1pt}\discretionary{.}{%
}{.}\hspace{.4pt}66}}}


\bibitem{related_event_schoning_2019}
J.~Sch{\"o}ning and G.~Heidemann.
\newblock Visual video analytics for interactive video content analysis.
\newblock In {\em Advances in Information and Communication Networks}, pp.
  346--360. Springer, Cham, Switzerland, 2019.
  \href{https://doi.org/10.1007/978-3-030-03402-3_23}
{doi: {{%
10\hspace{.1pt}\discretionary{.}{%
}{.}\hspace{.4pt}1007\discretionary{/}{%
}{/}978\discretionary{%
}{-}{-}3\discretionary{%
}{-}{-}030\discretionary{%
}{-}{-}03402\discretionary{%
}{-}{-}3\_23}}}


\bibitem{settles1994active}
B.~Settles.
\newblock Active learning literature survey.
\newblock {\em Machine Learning}, 15(2):201--221, 1994.

\bibitem{soomro2012ucf101}
K.~Soomro, A.~R. Zamir, and M.~Shah.
\newblock {UCF101}: A dataset of 101 human actions classes from videos in the
  wild.
\newblock {\em arXiv}, 2012. \href{https://doi.org/10.48550/arXiv.1212.0402}
{doi: {{%
10\hspace{.1pt}\discretionary{.}{%
}{.}\hspace{.4pt}48550\discretionary{/}{%
}{/}arXiv\hspace{.1pt}\discretionary{.}{%
}{.}\hspace{.4pt}1212\hspace{.1pt}\discretionary{.}{%
}{.}\hspace{.4pt}0402}}}


\bibitem{related_video_multimodal_soure_2022}
E.~J. Soure, E.~Kuang, M.~Fan, and J.~Zhao.
\newblock Coux: Collaborative visual analysis of think-aloud usability test
  videos for digital interfaces.
\newblock {\em IEEE Transactions on Visualization and Computer Graphics},
  28(1):643--653, 2022. \href{https://doi.org/10.1109/TVCG.2021.3114822}
{doi: {{%
10\hspace{.1pt}\discretionary{.}{%
}{.}\hspace{.4pt}1109\discretionary{/}{%
}{/}TVCG\hspace{.1pt}\discretionary{.}{%
}{.}\hspace{.4pt}2021\hspace{.1pt}\discretionary{.}{%
}{.}\hspace{.4pt}3114822}}}


\bibitem{sperrle2019viana}
F.~Sperrle, R.~Sevastjanova, R.~Kehlbeck, and M.~El-Assady.
\newblock Viana: Visual interactive annotation of argumentation.
\newblock In {\em Proc. VAST}, pp. 11--22. IEEE, Los Alamitos, 2019.
  \href{https://doi.org/10.1109/VAST47406.2019.8986917}
{doi: {{%
10\hspace{.1pt}\discretionary{.}{%
}{.}\hspace{.4pt}1109\discretionary{/}{%
}{/}VAST47406\hspace{.1pt}\discretionary{.}{%
}{.}\hspace{.4pt}2019\hspace{.1pt}\discretionary{.}{%
}{.}\hspace{.4pt}8986917}}}


\bibitem{sun2021task}
J.~Sun, A.~Kennedy, E.~Zhan, D.~Anderson, Y.~Yue, and P.~Perona.
\newblock Task programming: Learning data efficient behavior representations.
\newblock In {\em Proc. CVPR}, pp. 2875--2884. IEEE, Los Alamitos, 2021.
  \href{https://doi.org/10.1109/CVPR46437.2021.00290}
{doi: {{%
10\hspace{.1pt}\discretionary{.}{%
}{.}\hspace{.4pt}1109\discretionary{/}{%
}{/}CVPR46437\hspace{.1pt}\discretionary{.}{%
}{.}\hspace{.4pt}2021\hspace{.1pt}\discretionary{.}{%
}{.}\hspace{.4pt}00290}}}


\bibitem{related_event_tan_2022}
T.~Tang, Y.~Wu, Y.~Wu, L.~Yu, and Y.~Li.
\newblock Videomoderator: A risk-aware framework for multimodal video
  moderation in e-commerce.
\newblock {\em IEEE Transactions on Visualization and Computer Graphics},
  28(1):846--856, 2022. \href{https://doi.org/10.1109/TVCG.2021.3114781}
{doi: {{%
10\hspace{.1pt}\discretionary{.}{%
}{.}\hspace{.4pt}1109\discretionary{/}{%
}{/}TVCG\hspace{.1pt}\discretionary{.}{%
}{.}\hspace{.4pt}2021\hspace{.1pt}\discretionary{.}{%
}{.}\hspace{.4pt}3114781}}}


\bibitem{tsai-2019-multimodal}
Y.-H.~H. Tsai et~al.
\newblock Multimodal transformer for unaligned multimodal language sequences.
\newblock In {\em Proc. ACL}, pp. 6558--6569. ACL, Florence, Italy, 2019.
  \href{https://doi.org/10.18653/v1/P19-1656}
{doi: {{%
10\hspace{.1pt}\discretionary{.}{%
}{.}\hspace{.4pt}18653\discretionary{/}{%
}{/}v1\discretionary{/}{%
}{/}P19\discretionary{%
}{-}{-}1656}}}


\bibitem{vahdani2023deeplearning}
E.~Vahdani and Y.~Tian.
\newblock Deep learning-based action detection in untrimmed videos: A survey.
\newblock {\em IEEE Transactions on Pattern Analysis and Machine Intelligence},
  45(4):4302--4320, 2023. \href{https://doi.org/10.1109/TPAMI.2022.3193611}
{doi: {{%
10\hspace{.1pt}\discretionary{.}{%
}{.}\hspace{.4pt}1109\discretionary{/}{%
}{/}TPAMI\hspace{.1pt}\discretionary{.}{%
}{.}\hspace{.4pt}2022\hspace{.1pt}\discretionary{.}{%
}{.}\hspace{.4pt}3193611}}}


\bibitem{intro_wang_2021}
J.~Wang et~al.
\newblock Tac-valuer: Knowledge-based stroke evaluation in table tennis.
\newblock In {\em Proc. SIGKDD}, pp. 3688--3696. ACM, New York, 2021.
  \href{https://doi.org/10.1145/3447548.3467104}
{doi: {{%
10\hspace{.1pt}\discretionary{.}{%
}{.}\hspace{.4pt}1145\discretionary{/}{%
}{/}3447548\hspace{.1pt}\discretionary{.}{%
}{.}\hspace{.4pt}3467104}}}


\bibitem{wang2021actionclip}
M.~Wang, J.~Xing, and Y.~Liu.
\newblock Actionclip: A new paradigm for video action recognition.
\newblock {\em arXiv}, 2021. \href{https://doi.org/10.48550/arXiv.2109.08472}
{doi: {{%
10\hspace{.1pt}\discretionary{.}{%
}{.}\hspace{.4pt}48550\discretionary{/}{%
}{/}arXiv\hspace{.1pt}\discretionary{.}{%
}{.}\hspace{.4pt}2109\hspace{.1pt}\discretionary{.}{%
}{.}\hspace{.4pt}08472}}}


\bibitem{related_event_mm_wang_2022}
X.~Wang, J.~He, Z.~Jin, M.~Yang, Y.~Wang, and H.~Qu.
\newblock M2lens: Visualizing and explaining multimodal models for sentiment
  analysis.
\newblock {\em IEEE Transactions on Visualization and Computer Graphics},
  28(1):802--812, 2022. \href{https://doi.org/10.1109/TVCG.2021.3114794}
{doi: {{%
10\hspace{.1pt}\discretionary{.}{%
}{.}\hspace{.4pt}1109\discretionary{/}{%
}{/}TVCG\hspace{.1pt}\discretionary{.}{%
}{.}\hspace{.4pt}2021\hspace{.1pt}\discretionary{.}{%
}{.}\hspace{.4pt}3114794}}}


\bibitem{wang2022seq2pat}
X.~Wang, A.~Hosseininasab, P.~Colunga, S.~Kadıoğlu, and W.-J. van Hoeve.
\newblock Seq2pat: Sequence-to-pattern generation for constraint-based
  sequential pattern mining.
\newblock {\em Proc. AAAI}, 36:12665--12671, 2022.
  \href{https://doi.org/10.1609/aaai.v36i11.21542}
{doi: {{%
10\hspace{.1pt}\discretionary{.}{%
}{.}\hspace{.4pt}1609\discretionary{/}{%
}{/}aaai\hspace{.1pt}\discretionary{.}{%
}{.}\hspace{.4pt}v36i11\hspace{.1pt}\discretionary{.}{%
}{.}\hspace{.4pt}21542}}}


\bibitem{related_event_wang_2022}
X.~Wang, Y.~Ming, T.~Wu, H.~Zeng, Y.~Wang, and H.~Qu.
\newblock Dehumor: Visual analytics for decomposing humor.
\newblock {\em IEEE Transactions on Visualization and Computer Graphics},
  28(12):4609--4623, 2022. \href{https://doi.org/10.1109/TVCG.2021.3097709}
{doi: {{%
10\hspace{.1pt}\discretionary{.}{%
}{.}\hspace{.4pt}1109\discretionary{/}{%
}{/}TVCG\hspace{.1pt}\discretionary{.}{%
}{.}\hspace{.4pt}2021\hspace{.1pt}\discretionary{.}{%
}{.}\hspace{.4pt}3097709}}}


\bibitem{related_event_wang_2020}
X.~Wang, H.~Zeng, Y.~Wang, A.~Wu, Z.~Sun, X.~Ma, and H.~Qu.
\newblock Voicecoach: Interactive evidence-based training for voice modulation
  skills in public speaking.
\newblock In {\em Proc. CHI}, pp. 1--12. ACM, New York, 2020.
  \href{https://doi.org/10.1145/3313831.3376726}
{doi: {{%
10\hspace{.1pt}\discretionary{.}{%
}{.}\hspace{.4pt}1145\discretionary{/}{%
}{/}3313831\hspace{.1pt}\discretionary{.}{%
}{.}\hspace{.4pt}3376726}}}


\bibitem{wang2022historical}
Y.~Wang et~al.
\newblock Interactive visual exploration of longitudinal historical career
  mobility data.
\newblock {\em IEEE Transactions on Visualization and Computer Graphics},
  28(10):3441--3455, 2022. \href{https://doi.org/10.1109/TVCG.2021.3067200}
{doi: {{%
10\hspace{.1pt}\discretionary{.}{%
}{.}\hspace{.4pt}1109\discretionary{/}{%
}{/}TVCG\hspace{.1pt}\discretionary{.}{%
}{.}\hspace{.4pt}2021\hspace{.1pt}\discretionary{.}{%
}{.}\hspace{.4pt}3067200}}}


\bibitem{related_event_wong_2023}
K.~K. Wong, X.~Wang, Y.~Wang, J.~He, R.~Zhang, and H.~Qu.
\newblock Anchorage: Visual analysis of satisfaction in customer service videos
  via anchor events.
\newblock {\em IEEE Transactions on Visualization and Computer Graphics}, pp.
  1--13, 2023. \href{https://doi.org/10.1109/TVCG.2023.3245609}
{doi: {{%
10\hspace{.1pt}\discretionary{.}{%
}{.}\hspace{.4pt}1109\discretionary{/}{%
}{/}TVCG\hspace{.1pt}\discretionary{.}{%
}{.}\hspace{.4pt}2023\hspace{.1pt}\discretionary{.}{%
}{.}\hspace{.4pt}3245609}}}


\bibitem{related_event_wu_2020}
A.~Wu and H.~Qu.
\newblock Multimodal analysis of video collections: Visual exploration of
  presentation techniques in ted talks.
\newblock {\em IEEE Transactions on Visualization and Computer Graphics},
  26(7):2429--2442, 2020. \href{https://doi.org/10.1109/TVCG.2018.2889081}
{doi: {{%
10\hspace{.1pt}\discretionary{.}{%
}{.}\hspace{.4pt}1109\discretionary{/}{%
}{/}TVCG\hspace{.1pt}\discretionary{.}{%
}{.}\hspace{.4pt}2018\hspace{.1pt}\discretionary{.}{%
}{.}\hspace{.4pt}2889081}}}


\bibitem{intro_rasipam_2023}
J.~Wu, D.~Liu, Z.~Guo, and Y.~Wu.
\newblock Rasipam: Interactive pattern mining of multivariate event sequences
  in racket sports.
\newblock {\em IEEE Transactions on Visualization and Computer Graphics},
  29(1):940--950, 2023. \href{https://doi.org/10.1109/TVCG.2022.3209452}
{doi: {{%
10\hspace{.1pt}\discretionary{.}{%
}{.}\hspace{.4pt}1109\discretionary{/}{%
}{/}TVCG\hspace{.1pt}\discretionary{.}{%
}{.}\hspace{.4pt}2022\hspace{.1pt}\discretionary{.}{%
}{.}\hspace{.4pt}3209452}}}


\bibitem{yange2023example}
L.~Yang, C.~Xiong, J.~K. Wong, A.~Wu, and H.~Qu.
\newblock Explaining with examples: Lessons learned from crowdsourced
  introductory description of information visualizations.
\newblock {\em IEEE Transactions on Visualization and Computer Graphics},
  29(3):1638--1650, 2023. \href{https://doi.org/10.1109/TVCG.2021.3128157}
{doi: {{%
10\hspace{.1pt}\discretionary{.}{%
}{.}\hspace{.4pt}1109\discretionary{/}{%
}{/}TVCG\hspace{.1pt}\discretionary{.}{%
}{.}\hspace{.4pt}2021\hspace{.1pt}\discretionary{.}{%
}{.}\hspace{.4pt}3128157}}}


\bibitem{yang2022diagnosing}
W.~Yang et~al.
\newblock Diagnosing ensemble few-shot classifiers.
\newblock {\em IEEE Transactions on Visualization and Computer Graphics},
  28(9):3292--3306, 2022. \href{https://doi.org/10.1109/TVCG.2022.3182488}
{doi: {{%
10\hspace{.1pt}\discretionary{.}{%
}{.}\hspace{.4pt}1109\discretionary{/}{%
}{/}TVCG\hspace{.1pt}\discretionary{.}{%
}{.}\hspace{.4pt}2022\hspace{.1pt}\discretionary{.}{%
}{.}\hspace{.4pt}3182488}}}


\bibitem{yang2021induce}
Y.~Yang, J.~Kim, A.~Panagopoulou, M.~Yatskar, and C.~Callison-Burch.
\newblock Induce, edit, retrieve: Language grounded multimodal schema for
  instructional video retrieval.
\newblock {\em arXiv}, 2021. \href{https://doi.org/10.48550/arXiv.2111.09276}
{doi: {{%
10\hspace{.1pt}\discretionary{.}{%
}{.}\hspace{.4pt}48550\discretionary{/}{%
}{/}arXiv\hspace{.1pt}\discretionary{.}{%
}{.}\hspace{.4pt}2111\hspace{.1pt}\discretionary{.}{%
}{.}\hspace{.4pt}09276}}}


\bibitem{yuan2021survey}
J.~Yuan, C.~Chen, W.~Yang, M.~Liu, J.~Xia, and S.~Liu.
\newblock A survey of visual analytics techniques for machine learning.
\newblock {\em Computational Visual Media}, 7:3--36, 2021.
  \href{https://doi.org/10.1007/s41095-020-0191-7}
{doi: {{%
10\hspace{.1pt}\discretionary{.}{%
}{.}\hspace{.4pt}1007\discretionary{/}{%
}{/}s41095\discretionary{%
}{-}{-}020\discretionary{%
}{-}{-}0191\discretionary{%
}{-}{-}7}}}


\bibitem{related_event_zeng_2022}
H.~Zeng, X.~Wang, Y.~Wang, A.~Wu, T.-C. Pong, and H.~Qu.
\newblock Gesturelens: Visual analysis of gestures in presentation videos.
\newblock {\em IEEE Transactions on Visualization and Computer Graphics}, pp.
  3685--3697, 2022. \href{https://doi.org/10.1109/TVCG.2022.3169175}
{doi: {{%
10\hspace{.1pt}\discretionary{.}{%
}{.}\hspace{.4pt}1109\discretionary{/}{%
}{/}TVCG\hspace{.1pt}\discretionary{.}{%
}{.}\hspace{.4pt}2022\hspace{.1pt}\discretionary{.}{%
}{.}\hspace{.4pt}3169175}}}


\bibitem{related_event_zeng_2020}
H.~Zeng, X.~Wang, A.~Wu, Y.~Wang, Q.~Li, A.~Endert, and H.~Qu.
\newblock Emoco: Visual analysis of emotion coherence in presentation videos.
\newblock {\em IEEE Transactions on Visualization and Computer Graphics},
  26(1):927--937, 2020. \href{https://doi.org/10.1109/TVCG.2019.2934656}
{doi: {{%
10\hspace{.1pt}\discretionary{.}{%
}{.}\hspace{.4pt}1109\discretionary{/}{%
}{/}TVCG\hspace{.1pt}\discretionary{.}{%
}{.}\hspace{.4pt}2019\hspace{.1pt}\discretionary{.}{%
}{.}\hspace{.4pt}2934656}}}


\bibitem{zhang2022survey}
J.~Zhang, C.~Hsieh, Y.~Yu, C.~Zhang, and A.~Ratner.
\newblock A survey on programmatic weak supervision.
\newblock {\em arXiv}, 2022. \href{https://doi.org/10.48550/ARXIV.2202.05433}
{doi: {{%
10\hspace{.1pt}\discretionary{.}{%
}{.}\hspace{.4pt}48550\discretionary{/}{%
}{/}ARXIV\hspace{.1pt}\discretionary{.}{%
}{.}\hspace{.4pt}2202\hspace{.1pt}\discretionary{.}{%
}{.}\hspace{.4pt}05433}}}


\bibitem{wong_cohortva_2023}
W.~Zhang et~al.
\newblock Cohortva: A visual analytic system for interactive exploration of
  cohorts based on historical data.
\newblock {\em IEEE Transactions on Visualization and Computer Graphics},
  29(1):756--766, 2023. \href{https://doi.org/10.1109/TVCG.2022.3209483}
{doi: {{%
10\hspace{.1pt}\discretionary{.}{%
}{.}\hspace{.4pt}1109\discretionary{/}{%
}{/}TVCG\hspace{.1pt}\discretionary{.}{%
}{.}\hspace{.4pt}2022\hspace{.1pt}\discretionary{.}{%
}{.}\hspace{.4pt}3209483}}}


\bibitem{zhang2022onelabeler}
Y.~Zhang, Y.~Wang, H.~Zhang, B.~Zhu, S.~Chen, and D.~Zhang.
\newblock Onelabeler: A flexible system for building data labeling tools.
\newblock In {\em Proc. CHI}, pp. 1--22. ACM, New York, 2022.
  \href{https://doi.org/10.1145/3491102.3517612}
{doi: {{%
10\hspace{.1pt}\discretionary{.}{%
}{.}\hspace{.4pt}1145\discretionary{/}{%
}{/}3491102\hspace{.1pt}\discretionary{.}{%
}{.}\hspace{.4pt}3517612}}}


\bibitem{wong_dpviscreator_2023}
J.~Zhou et~al.
\newblock Dpviscreator: Incorporating pattern constraints to privacy-preserving
  visualizations via differential privacy.
\newblock {\em IEEE Transactions on Visualization and Computer Graphics},
  29(1):809--819, 2023. \href{https://doi.org/10.1109/TVCG.2022.3209391}
{doi: {{%
10\hspace{.1pt}\discretionary{.}{%
}{.}\hspace{.4pt}1109\discretionary{/}{%
}{/}TVCG\hspace{.1pt}\discretionary{.}{%
}{.}\hspace{.4pt}2022\hspace{.1pt}\discretionary{.}{%
}{.}\hspace{.4pt}3209391}}}


\end{thebibliography}


\appendix 







\end{document}